%% file: main.tex
\documentclass[journal]{IEEEtran}
\usepackage{amsmath,amsfonts}
\usepackage{algorithmic}
\usepackage{algorithm}
\usepackage{array}
\usepackage[caption=false,font=normalsize,labelfont=sf,textfont=sf]{subfig}
\usepackage{textcomp}
\usepackage{stfloats}
\usepackage{url}
\usepackage{verbatim}
\usepackage{graphicx}
\usepackage{cite}
\usepackage{paralist}
\usepackage{subcaption}
\usepackage{booktabs}
\usepackage{multirow}
\usepackage{pifont}
\usepackage[dvipsnames]{xcolor}

\def\ie{\emph{i.e.}}
\def\eg{\emph{e.g.}}
\def\vs{\emph{vs.~}}
\def\etc{\emph{etc}}

\definecolor{rblue}{rgb}{0,0.5,1}
\definecolor{hollywoodcerise}{rgb}{0.96, 0.0, 0.63}
\definecolor{lasallegreen}{rgb}{0.03, 0.47, 0.19}
\definecolor{hanpurple}{rgb}{0.32, 0.09, 0.98}
\definecolor{green(pigment)}{rgb}{0.0, 0.65, 0.31}

\definecolor{mygreen}{RGB}{20, 180, 70}
\definecolor{myorange}{RGB}{255, 140, 0}
\definecolor{myred}{RGB}{220, 30, 30}
\definecolor{mygray}{RGB}{211, 211, 211}
\definecolor{mybule}{RGB}{39, 125, 142}

\newcommand{\cmark}{{\color{mygreen}\ding{51}}}
\newcommand{\cmarkOrange}{{\color{myorange}(\ding{51})}}
\newcommand{\xmark}{{\color{myred}\ding{56}}}

\definecolor{rblue}{rgb}{0,0.5,1}

\usepackage[pagebackref=false,breaklinks=true,colorlinks,bookmarks=false]{hyperref}
\hypersetup{colorlinks,linkcolor={red},citecolor={hanpurple},urlcolor={magenta}}  

\begin{document}

\title{Panoramic Out-of-Distribution Segmentation}

\author{Mengfei Duan, Yuheng Zhang, Yihong Cao, Fei Teng, Kai Luo, Jiaming Zhang,\\Kailun Yang,~\IEEEmembership{Member,~IEEE}, and Zhiyong Li,~\IEEEmembership{Member,~IEEE}
\thanks{This work was supported in part by the National Natural Science Foundation of China (Grant No. 62388101, No. U23A20341, and No. 62473139), in part by the Major Scientific and Technological Research Plan of Hunan Province (No. 2025QK1005), in part by the Hunan Provincial Research and Development Project (Grant No. 2025QK3019), and in part by the State Key Laboratory of Autonomous Intelligent Unmanned Systems (the opening project number ZZKF2025-2-10).
\textit{(Corresponding authors: Kailun Yang and Zhiyong Li.)}}
\thanks{M. Duan, Y. Zhang, F. Teng, K. Luo, J. Zhang, K. Yang, and Z. Li are with the School of Artificial Intelligence and Robotics and the National Engineering Research Center of Robot Visual Perception and Control Technology, Hunan University, Changsha 410082, China (e-mail: kailun.yang@hnu.edu.cn; zhiyong.li@hnu.edu.cn).}
\thanks{Y. Cao is with the Key Laboratory of Big Data Research and Application for Basic Education, Hunan Normal University, Changsha 410006, China.}
}

\markboth{IEEE Transactions on Circuits and Systems for Video Technology, October~2026}%
{Duan~\MakeLowercase{\textit{et al.}}: Panoramic Out-of-Distribution Segmentation}

\maketitle

\input{Tex_content/abstract}

\begin{IEEEkeywords}
Scene Segmentation, Panoramic Images, Out-of-Distribution Segmentation, Scene Understanding.
\end{IEEEkeywords}

\input{Tex_content/introduction}
\input{Tex_content/related_work}
\input{Tex_content/method}
\input{Tex_content/benchmark}
\input{Tex_content/experiments}
\input{Tex_content/conclusion}

\bibliographystyle{IEEEtran}
\bibliography{bib}

\input{Tex_content/appendix}

\end{document}

%% file: Tex_content/abstract.tex
\begin{abstract}
Panoramic imaging enables capturing $360^{\circ}$ images with an ultra-wide Field-of-View (FoV) for dense omnidirectional perception, which is critical to applications such as autonomous driving and augmented reality. However, current panoramic semantic segmentation methods fail to identify outliers, and pinhole out-of-distribution segmentation (pinhole-OoS) models perform unsatisfactorily in the panoramic domain due to pixel distortions and background clutter. To address these issues, we introduce a new task, Panoramic Out-of-Distribution Segmentation (PanOoS), with the aim of achieving comprehensive and safe scene understanding. Furthermore, we propose the first solution, POS, which adapts to the characteristics of panoramic images through text-guided prompt distribution learning. Specifically, POS integrates a disentanglement strategy designed to materialize the cross-domain generalization capability of CLIP. The proposed Prompt-based Restoration Attention (PRA) optimizes semantic decoding by prompt guidance and self-adaptive correction, while Bilevel Prompt Distribution Learning (BPDL) refines the manifold of per-pixel mask embeddings via semantic prototype supervision. Besides, to compensate for the scarcity of PanOoS datasets, we establish two benchmarks: DenseOoS, which features diverse outliers in complex environments, and QuadOoS, captured by a quadruped robot with a panoramic annular lens system. Extensive experiments demonstrate the effectiveness of POS, with AuPRC increasing and FPR$_{95}$ decreasing by $34.25$ and $21.42$ percentage points, respectively, compared with RbA on DenseOoS, outperforming state-of-the-art pinhole-OoS methods. Moreover, POS achieves leading closed-set segmentation capabilities and advances the development of panoramic understanding. Code and datasets will be available at~\href{https://github.com/MengfeiD/PanOoS}{PanOoS}.
\end{abstract}

%% file: Tex_content/introduction.tex
\section{Introduction}
\label{sec:intro}
\IEEEPARstart{O}{mnidirectional} perception technologies are vital for embodied agents, such as autonomous vehicles and mobile robots, enabling comprehensive understanding of the surroundings~\cite{zhang2024behind,ai2025survey}. In particular, panoramic imaging technology, capable of capturing high-quality $360^{\circ}$ images via optical systems, has gained significant attention across various omnidirectional vision tasks~\cite{zheng2025one_flight}, such as dense visual prediction~\cite{shen2022panoformer,ling2023panoswin,yu2023panelnet}, holistic scene understanding~\cite{zhang2021deeppanocontext,zheng2025open}, and panoramic scene segmentation~\cite{yang2020omnisupervised,zhang2022bending,zheng2023both,cao2024occlusion}. 
However, existing panoramic semantic segmentation models are trained to recognize a predefined set of semantic classes (\eg, \emph{road}, \emph{building}, \emph{car}, \emph{pedestrian})~\cite{cordts2016cityscapes,ma2021densepass,huang2024high}, and thus fail to identify rare, long-tail, or semantically ambiguous objects that do not belong to the predefined classes, \eg, \emph{animals} or \emph{furniture}. The broader Field-of-View (FoV) of panoramic images further increases the likelihood of capturing such non-predefined objects, thereby exacerbating the outlier identification problem. This remains a persistent challenge in deploying robust perception systems and other safety-critical applications, where environments are inherently dynamic, diverse, and unpredictable. Therefore, there is an urgent need for a technology capable of identifying outliers not present during training to avoid potential dangers.

\begin{figure}[!t]
    \centering
    \includegraphics[width=.93\linewidth]{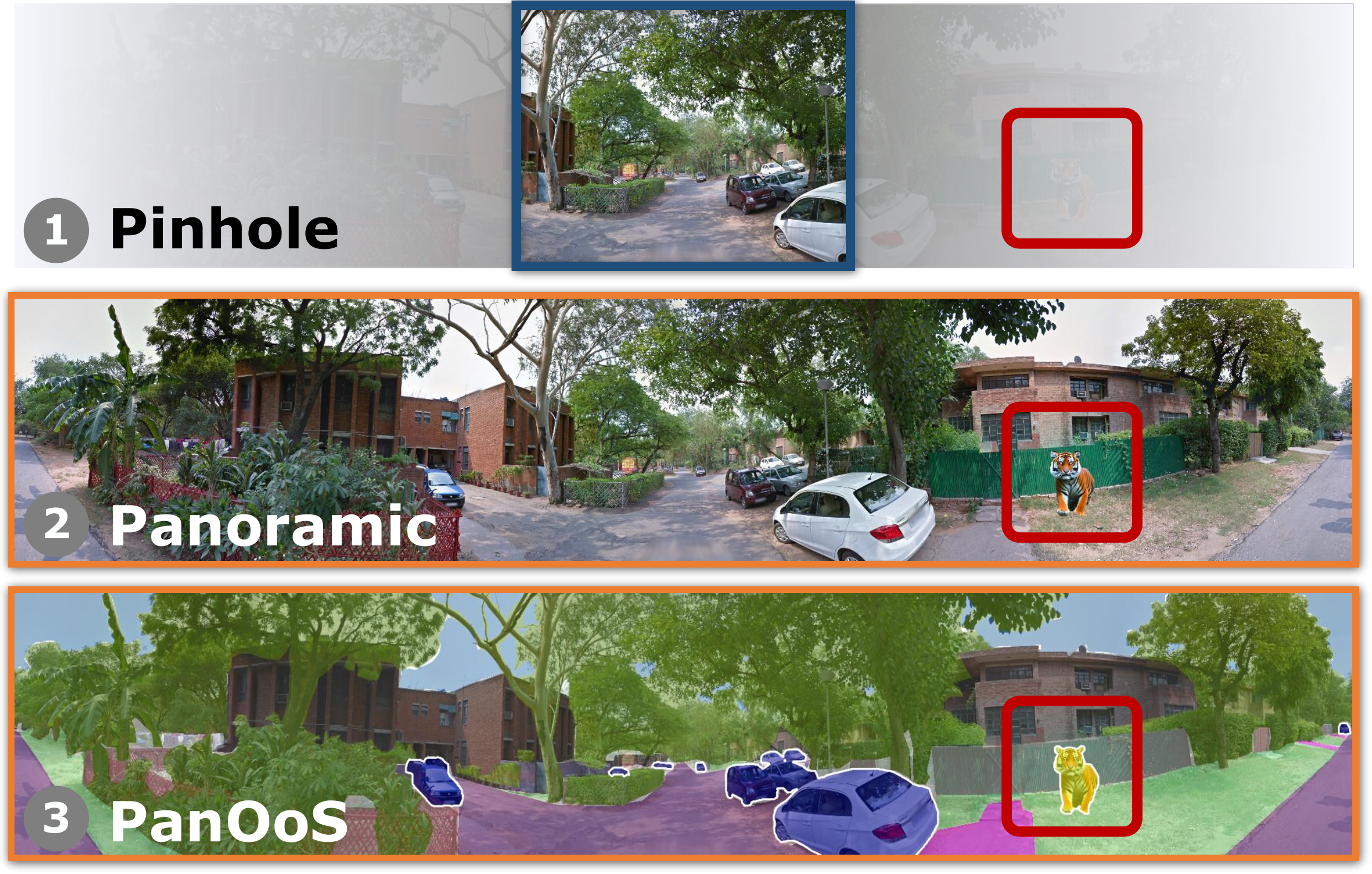}
    \caption{\textbf{\emph{PanOoS}} enhances comprehensive and safe omnidirectional scene perception. 
    \normalsize{\textcircled{\scriptsize{1}}}\normalsize: pinhole image with narrow FoV; 
    \normalsize{\textcircled{\scriptsize{2}}}\normalsize: $360^{\circ}$ panoramic image; 
    \normalsize{\textcircled{\scriptsize{3}}}\normalsize: segmentation result. 
    The \emph{tiger} in the image is an outlier on the street.}
    \label{fig:panOoS}
    \vskip -2ex
\end{figure}

To enhance comprehensive and safe omnidirectional scene understanding, we introduce a new task, termed \emph{Panoramic Out-of-Distribution Segmentation (PanOoS)}. 
Unlike pinhole out-of-distribution segmentation (pinhole-\emph{OoS}), \emph{PanOoS} offers a broader FoV beyond the pinhole view, for example, ranging from $95^{\circ}$ to $360^{\circ}$ as depicted in Fig.~\ref{fig:panOoS}, enabling detection of outliers and even dangers not visible from a pinhole perspective, \eg, a \emph{tiger}. 
Yet, panoramic images pose unique challenges in visual understanding tasks due to their broader FoV and complex scene content, which introduce pixel distortions~\cite{gao2022review,yang2021capturing} and background clutter, thereby complicating semantic interpretation. 
These characteristics lead to feature distribution shifts and loss of semantic information, impairing the model’s ability to capture low-level features and fine details of the image. Moreover, these issues interfere with foreground object recognition, introduce semantic ambiguity in overlapping or visually similar regions, and complicate distinction between in-distribution and out-of-distribution (OoD) areas.

\begin{figure}[!t]
    \centering
    \includegraphics[width=\linewidth]{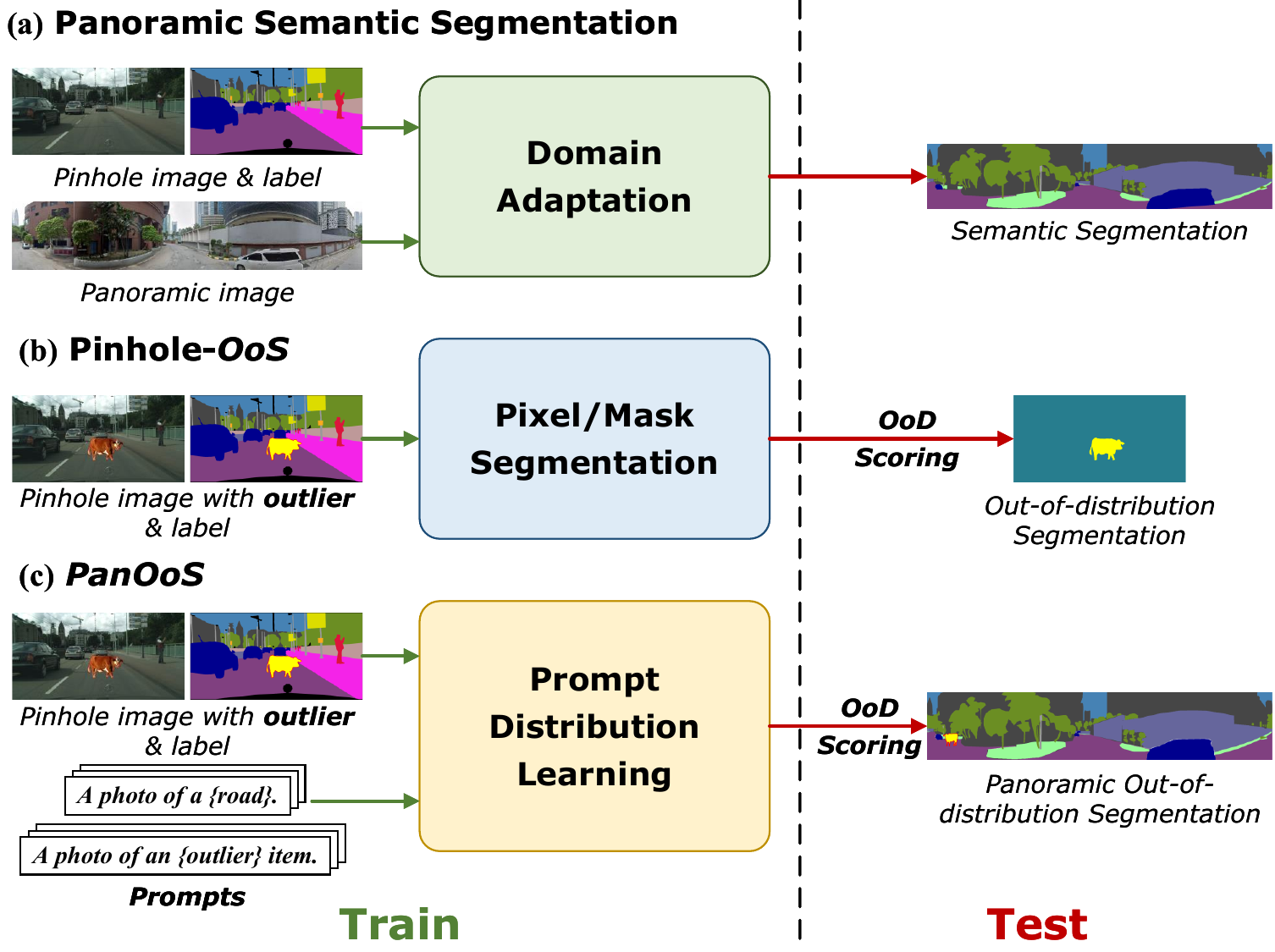}
    \vskip -1ex
    \caption{\textbf{Comparison of panoramic semantic and out-of-distribution segmentation paradigms.} 
    (a) Panoramic semantic segmentation methods adapt the model from pinhole to panoramas through domain adaptation, yet remain limited to in-distribution dense predictions; 
    (b) Pinhole-\emph{OoS} methods combine semantic segmentation with OoD scoring to detect outliers, but are restricted to narrow pinhole FoV; 
    (c) POS achieves segmentation in the distribution space of panoramic images via prompt distribution learning. 
    The pinhole images denote the training-domain samples, including labeled inlier data and auxiliary outlier-exposure data, while the panoramic images denote the target-domain evaluation data.}
    \label{fig:architectureDif}
    \vskip -2ex
\end{figure}

\begin{figure*}[!t]
    \centering
    \includegraphics[width=.9\textwidth]{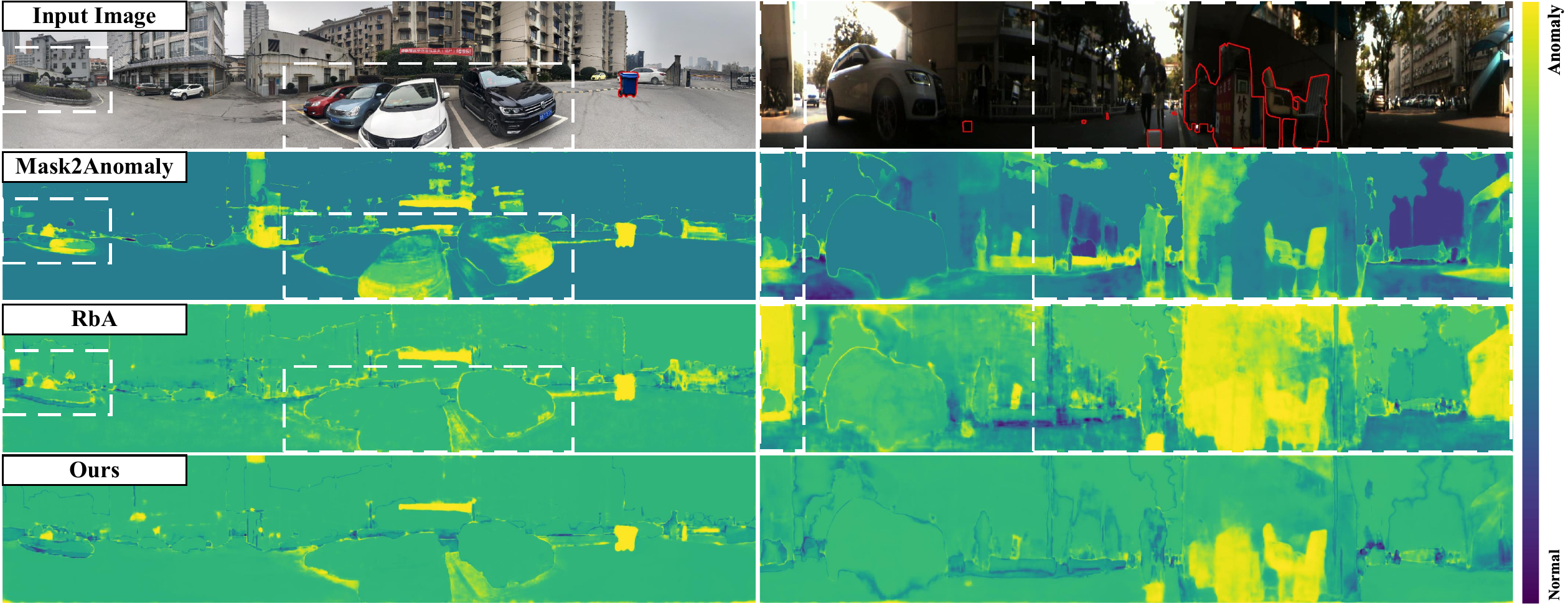}
    \caption{\textbf{Impact of pixel distortions and background clutter} on state-of-the-art pinhole-\emph{OoS} methods. 
    The white dashed boxes highlight representative challenging regions, while the red solid contours indicate anomalous objects or regions. 
    Existing pinhole-\emph{OoS} methods, \eg, Mask2Anomaly and RbA, fail to adapt to pixel distortions and background clutter, resulting in a higher number of false positives. In contrast, the proposed POS effectively enhances semantic perception of panoramic pixels.}
    \label{fig:problemVis}
    \vskip -2ex
\end{figure*}

Although several effective techniques exist for panoramic semantic segmentation~\cite{zhang2022bending,zheng2023look,li2023sgat4pass}, they are only designed for closed-set segmentation. 
The inability to handle outliers means that these techniques may miss critical objects, leading to incorrect or incomplete scene understanding. 
This issue limits their practical applicability, particularly in dynamic and unpredictable environments where outliers frequently occur. 
High costs associated with data annotation have also resulted in a scarcity of datasets specifically designed for training panoramic segmentation models. 
As illustrated in Fig.~\ref{fig:architectureDif}, most panoramic segmentation methods rely on techniques such as unsupervised domain adaptation~\cite{zheng2023both,zheng2023look,cao2024occlusion} or knowledge adaptation~\cite{zhang2024goodsam,jaus2023panoramic_insights,zhang2024goodsam++} to transfer models from the pinhole to the panoramic imagery. 
However, these approaches are not suitable for \emph{PanOoS}, where the model must not only adapt to the characteristics of the panoramic imagery, but also learn to distinguish between in-distribution regions and outliers while ensuring closed-set segmentation performance. 
Meanwhile, state-of-the-art pinhole-\emph{OoS} methods~\cite{grcic2023advantages,nayal2023rba,rai2024mask2anomaly} for pinhole images typically involve first training a mask-based segmentation model, \eg, in Mask2Former style~\cite{cheng2022masked}, and then computing OoD scores using various outlier detection techniques during fine-tuning or inference. 
Nevertheless, existing research works primarily focus on narrow-FoV pinhole images, and thus lack the architectural or learning mechanisms needed to handle the geometric distortions and wide contextual dependencies inherent in panoramic imagery, making them ill-equipped for \emph{PanOoS} (see Fig.~\ref{fig:problemVis}).

To tackle the challenges of \emph{PanOoS}, we propose POS, a mask-based framework that integrates prompt distribution learning to enable accurate pixel-wise detection of OoD regions in complex and distorted panoramic scenes. 
(1) Compared to purely visual models, vision-language models are typically exposed to a broader array of visual concepts during pre-training~\cite{yu2023convolutions,zhang2024language,zhao2024focusclip}. 
To enhance the pixel-level perception of outliers within panoramic scenes, POS incorporates CLIP~\cite{radford2021learning} into a mask-transformer architecture, leveraging its strong generalization capabilities to better capture scene semantics. Moreover, during outlier exposure fine-tuning, POS explicitly disentangles per-pixel mask decoding from semantic classification, thereby mitigating semantic entanglement and preserving accurate mask representations for subsequent pixel-level decoding. 
(2) Traditional image-based query decoding methods~\cite{cheng2022masked,grcic2023advantages,nayal2023rba,rai2024mask2anomaly} often struggle to accurately capture semantic information under pixel distortions and background clutter. To address this, we propose Prompt-based Restoration Attention (PRA), which leverages the text encoder to extract semantic information from prompts at both class-level and distributional levels. By adaptively guiding the learning process, the PRA mechanism optimizes semantic extraction, enhancing the accuracy and robustness of semantic understanding, thereby mitigating semantic ambiguity in challenging panoramic scenes. 
(3) Finally, a Bilevel Prompt Distribution Learning (BPDL) mechanism is proposed. By leveraging the language-image semantic distribution consistency of the CLIP model, BPDL refines the distribution of per-pixel mask embeddings and strengthens the distinguishability of out-of-distribution features and the adaptability of the pinhole-trained model to severe distortions encountered in panoramic images. 

With these crucial designs, where the panoramic-oriented disentanglement strategy preserves reliable mask decoding during outlier-exposure fine-tuning, PRA restores query-level semantic representations, and BPDL regularizes the pixel-level embedding distribution, POS achieves superior alignment with the characteristics of panoramic images, enabling efficient out-of-distribution segmentation while demonstrating impressive closed-set segmentation performance.

To facilitate the evaluation of the \emph{PanOoS} task, we create two datasets: 
(1) We build the first panoramic out-of-distribution segmentation dataset, Dense Panoramic Out-of-Distribution Segmentation (DenseOoS). 
Due to the broader FoV, the background of each panoramic image is highly complex, featuring object instances that vary in shape and appearance, along with significant distortions. 
Meanwhile, outliers are placed in semantically plausible regions to ensure realism, for which we spent significant effort on manual screening. 
(2) To further verify the performance of POS in practical application and explore \emph{PanOoS} in unconstrained, unstructured, and intense-motion real-world scenarios, we employ a quadruped mobile robot to collect panoramic images containing outliers, creating the Quadruped Panoramic Out-of-Distribution Segmentation (QuadOoS) dataset. 
Compared to DenseOoS, QuadOoS contains a greater number of outliers per image, with smaller sizes, while maintaining similar panoramic imaging characteristics. 
Ultimately, based on DenseOoS and QuadOoS, we establish the evaluation benchmarks for the \emph{PanOoS} task. The proposed POS achieves state-of-the-art performance on both benchmarks, outperforming other pinhole-\emph{OoS} methods~\cite{tian2022pixel,grcic2022densehybrid,rai2024mask2anomaly,nayal2023rba,grcic2023advantages}, with AuPRC, and FPR$_{95}$ of $(85.56\%, 48.13\%)$, and $(0.45\%, 85.22\%)$, respectively. 
Furthermore, compared to panoramic semantic segmentation methods~\cite{hoyer2022daformer,zhang2022bending,zhang2024behind}, it demonstrates superior closed-set segmentation performance, with an mIoU of $63.30\%$ on DenseOoS. 
These results collectively validate the effectiveness of the proposed POS in addressing \emph{PanOoS} and enhancing panoramic scene understanding.

At a glance, this work delivers the following contributions:
\begin{compactitem}
    \item We introduce the Panoramic Out-of-Distribution Segmentation (\emph{PanOoS}) task, which, for the first time, explores out-of-distribution segmentation in panoramic imagery, aiming to facilitate holistic scene understanding. 
    \item To achieve the \emph{PanOoS} task, we propose \emph{POS}, a mask-transformer-based model with Prompt-based Restoration Attention (PRA) and Bilevel Prompt Distribution Learning (BPDL), which effectively alleviates the effects of pixel distortions and background clutter, demonstrating strong performance. 
    \item Two datasets, DenseOoS and QuadOoS, are created, serving as important benchmarks for \emph{PanOoS} and contributing to the advancement of panoramic understanding.
\end{compactitem}

%% file: Tex_content/related_work.tex
\section{Related Work}
\label{sec:relatedWork}

\subsection{Panoramic Semantic Segmentation}
By capturing broader FoV scenes, panoramic images serve as a starting point for a more holistic understanding of $360^{\circ}$ surroundings~\cite{yang2019pass,jiang2024minimalist}. 
The first line of research on panoramic semantic segmentation~\cite{xu2019semantic,orhan2022semantic_outdoor,li2023sgat4pass,kim2022pasts,guttikonda2024single} is based on supervised learning. 
However, to address the scarcity of annotations, researchers have revisited panoramic segmentation from the perspective of unsupervised domain adaptation~\cite{jang2022dada,zhu2023patch,zhang2021transfer,jiang2025gaussian,yuan2023laformer} by leveraging rich training sources from narrow-FoV pinhole data. 
The approaches can be divided into three types: pseudo-labeling~\cite{liu2021pano,zhang2021deeppanocontext}, adversarial training~\cite{ma2021densepass,zheng2023both,jiang2025multi}, and prototype adaptation~\cite{zhang2024behind,zheng2024semantics,cao2024occlusion,zheng2024360sfuda++,zhong2025omnisam}. 
Pseudo-labeling generates self-supervised labels for the panoramic domain. 
Adversarial training encourages the target model to capture domain-invariant characteristics across domains using a discriminator. 
Prototype adaptation approaches align high-level feature centers between the pinhole and panoramic domains. 
Additionally, related work demonstrates strong performance from the perspective of knowledge adaptation~\cite{zhang2024goodsam,zhang2024goodsam++,jaus2023panoramic_insights}. 
However, existing methods focus on closed-set segmentation and do not address the problem of identifying outliers within the scene. 
Recently, open panoramic segmentation~\cite{zheng2025open} has been proposed, but the focus has been on improving the efficiency of open-vocabulary segmentation. 
In contrast to prior work, we aim to explore out-of-distribution segmentation in panoramic images and leverage the distribution learning of text-based prompts to advance panoramic scene understanding.

\subsection{Out-of-distribution Segmentation}
Out-of-distribution segmentation (OoS) aims to identify and separate outliers in specific scenes, such as complex driving environments~\cite{chan2021segmentmeifyoucan,blum2021fishyscapes,cen2023consensus,wan2025out}, while maintaining closed-set segmentation capability. 
Early uncertainty-based methods assume that outliers lead to low-confidence predictions and focus on estimating the prediction uncertainty through various approaches to measure pixel-wise anomaly scores~\cite{mukhoti2018evaluating,corbiere2019addressing,hendrycks2019scaling,jung2021standardized,cen2021deep}. However, models trained on closed sets may make high-confidence erroneous predictions for unseen categories. Subsequent works have shifted their focus to reconstruction. Reconstruction-based methods detect outliers by measuring the discrepancy between the original image and a reconstruction generated from semantic segmentation predictions~\cite{lis2019detecting,xia2020synthesize,haldimann2019not,vojir2021road,grcic2021dense,lis2023detecting}. However, nearly all reconstruction-based methods are highly sensitive to the quality of the reconstructed output, which can lead to performance degradation due to image artifacts~\cite{fontanel2021detecting}. 
These methods also suffer from context dependence and high computational overhead, resulting in unsatisfactory performance in complex environments.

Recent approaches~\cite{chan2021entropy,tian2022pixel,grcic2022densehybrid,liu2023residual,choi2023balanced} leverage the Outlier Exposure (OE) strategy~\cite{hendrycks2018deep} that utilizes auxiliary out-of-distribution data to regularize the model's feature space. These auxiliary outliers are typically images from other datasets (such as ImageNet~\cite{deng2009imagenet}, COCO~\cite{lin2014microsoft}, ADE20K~\cite{zhou2017scene}, \etc.) or synthetic composites created by cutting out anomalous objects from external images and pasting them into inlier scenes. 
Notably, PEBAL~\cite{tian2022pixel} learns energy-based adaptive penalties through abstention learning, whereas DenseHybrid~\cite{grcic2022densehybrid} combines likelihood and posterior evaluation to achieve better results. 
Although the OE strategy generally improves performance over earlier approaches, pixel-wise architectures that adopt it still score each pixel independently without considering local semantic context, leading to noisy anomaly predictions. 
Currently, mask-transformer-based methods~\cite{ackermann2023maskomaly,grcic2023advantages,rai2024mask2anomaly,nayal2023rba,delic2024outlier,zhang2024csl} overcome this limitation by segmenting outliers into semantically clustered masks and encouraging the objectness of predictions, achieving leading segmentation results. 
EAM~\cite{grcic2023advantages} ensembles region-based outlier scores, whereas Mask2Anomaly~\cite{rai2024mask2anomaly} combines contrastive learning and mask refinement to significantly reduce false positives. 
RbA~\cite{nayal2023rba} utilizes mask queries that act as a one-vs-all classifier~\cite{grcic2023advantages,nayal2023rba,shoeb2025out}, helping to eliminate uncertainty in ambiguous in-point regions, \eg, semantic boundaries, by reducing the scores in negative pixels. 
Despite these advances, existing methods primarily focus on narrow-FoV pinhole perception and parsing granularity~\cite{mohan2024panoptic}, and when transferred to the panoramic domain, they struggle to adapt to challenges such as pixel distortions and background clutter caused by the broad FoV. 
To the best of our knowledge, this is the first work to tackle this pressing problem of out-of-distribution segmentation in panoramic imagery.

%% file: Tex_content/method.tex
\section{Methodology}
\label{sec:method}
In this work, we introduce a novel panoramic perception task, \emph{PanOoS}, to promote a more comprehensive omnidirectional perception by leveraging the broader FoV of panoramic imagery. In response to the challenges posed by pixel distortions and complex backgrounds, we analyze the model architecture and fine-tuning strategies, ultimately proposing the first mask-based model tailored to the \emph{PanOoS} task.

\subsection{Problem Setting}
Out-of-distribution segmentation generalizes anomaly segmentation by requiring not only the detection of anomalous pixels but also accurate semantic labeling of all in-distribution categories. Formally, given an input image $\mathbf{x}{\in}\mathbb{R}^{3\times H \times W}$, the pixel-wise segmentation architectures~\cite{cheng2022masked,nayal2023rba,rai2024mask2anomaly,grcic2023advantages,delic2024outlier} produce per-pixel class logits $\mathbf{S}(x){\in}\mathbb{R}^{K\times H\times W}$, where $K$ denotes the number of classes for closed-set segmentation, and $H$ and $W$ correspond to the height and width of the image, respectively. An anomaly score map $\mathbf{A}(x)$ is defined as
\begin{equation}
  \mathbf{A}(x) = f(\mathbf{S}(x)) \in \mathbb{R}^{H \times W},
\end{equation}
where $f{:}~\mathbb{R}^{K} \to \mathbb{R}$ maps the per-pixel class logits to a scalar out-of-distribution confidence score.

\begin{figure*}[!t]
    \centering
    \includegraphics[width=\textwidth]{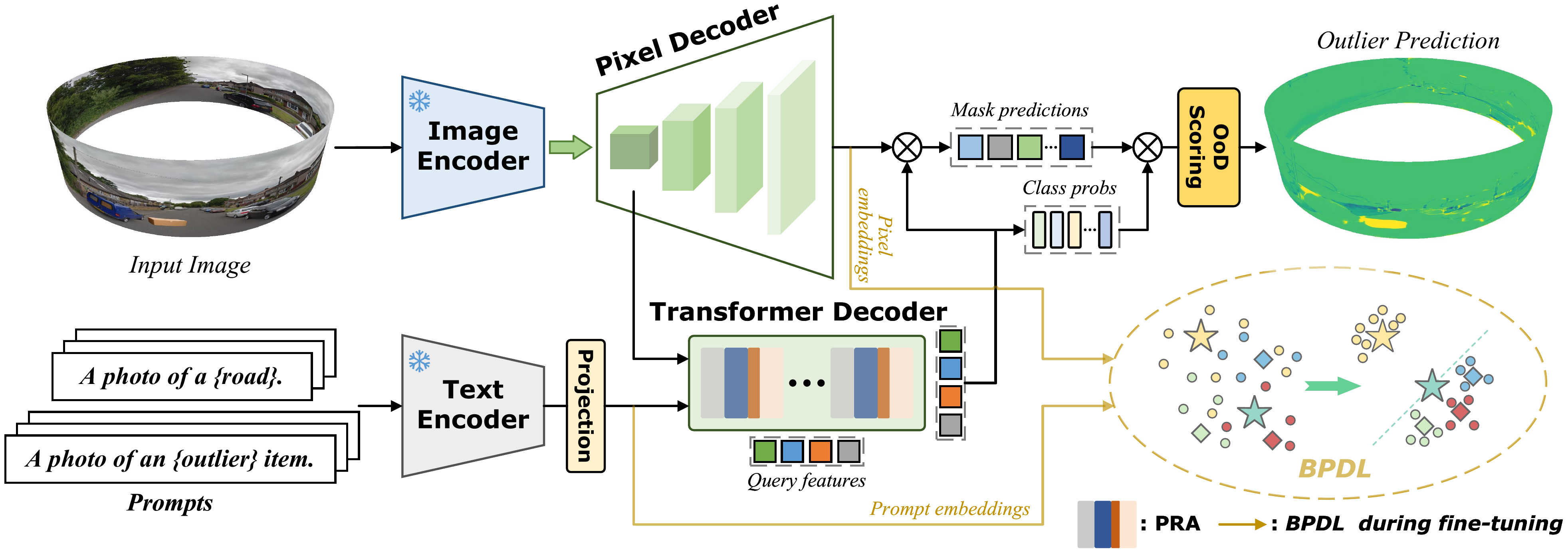}
    \vskip -1ex
    \caption{
    \textbf{Overview of POS.} POS consists of an image encoder, a text encoder, a pixel decoder, and a Transformer decoder equipped with Prompt-based Restoration Attention~(PRA). During outlier exposure fine-tuning, Bilevel Prompt Distribution Learning~(BPDL) regularizes the embedding manifold of per-pixel mask embeddings with class-level and distribution-level prompt priors. See Fig.~\ref{fig:bpdl} for details of BPDL. During inference, POS jointly processes image features and prompt embeddings to predict masks and class probabilities, which are further integrated into per-pixel class logits for final OoD scoring.
    }
    \label{fig:pos}
    \vskip -2ex
\end{figure*}

\subsection{Framework}
Despite the strong performance of the mask-transformer model~\cite{cheng2022masked} in pinhole-\emph{OoS}, the current methods fail to generalize to \emph{PanOoS}. The domain adaptation barrier prevents the model from effectively learning the visual features of panoramic images, resulting in suboptimal segmentation performance. It is important to explore methods for transferring models trained in the pinhole domain to the panoramic domain. Moreover, accurate pixel decoding is essential for panoramic segmentation. During the critical outlier exposure fine-tuning, a conflict arises between the query embedding, which favors semantic classification, and the pixel decoding, which focuses on image details. Therefore, we optimize the architecture and fine-tuning procedures, introducing POS, the first mask-transformer-based segmentation model for \emph{PanOoS}, as illustrated in Fig.~\ref{fig:pos}. The model comprises four main components: the image encoder, the text encoder, the pixel decoder, and the Transformer decoder.

\subsubsection{Image Encoder}
We utilize the image encoder from CLIP~\cite{radford2021learning} for extracting image features and freeze its weights. The image encoder is a ConvNeXt-Large model pretrained in the CLIP framework~\cite{liu2022convnet}, comprising four stages. It accepts an RGB image $\mathbf{x} \in \mathbb{R}^{3\times H\times W}$ as input and outputs feature maps at several resolutions to the pixel decoder. Specifically, the output feature maps are downsampled with strides of $4 (\mathbf{c}_{4})$, $8 (\mathbf{c}_{8})$, $16 (\mathbf{c}_{16})$, and $32 (\mathbf{c}_{32})$ relative to the input image.

\subsubsection{Text Encoder}
For the textual features, the $K$ semantic class names and distribution nouns, \emph{inlier} and \emph{outlier}, are integrated into the pre-designed prompt templates to obtain text descriptions about classes and distributions. The text encoder is structured as a $16$-layer transformer, each layer being $768$ units wide and featuring $12$ attention heads. The descriptions are fed into the text encoder to generate the prompt embeddings $\{\mathbf{T}_{i} \mid i=0,\dots,K\} \cup \{\mathbf{P}_{in}, \mathbf{P}_{out}\} {\in} \mathbb{R}^{(K+3)\times768}$, where $\mathbf{T}_{0}$ denotes the learnable void embedding. A linear projection is then applied to project the prompt embeddings into the pixel space to match the dimensionality of per-pixel embeddings. Prompt template details are in the Appendix.

\subsubsection{Pixel Decoder}
Following Mask2Former~\cite{cheng2022masked}, the pixel decoder comprises $6$ layers of Deformable Attention Transformer~(DAT)~\cite{zhu2020deformable} and a Feature Pyramid Network~(FPN)~\cite{lin2017feature}. 
To facilitate \emph{OoS}, the enriched semantic feature map $\mathbf{c}_{32}$ is processed by the DAT layers to produce $\mathbf{f}_{4}$, which is then passed to the transformer decoder. 
Then, the pixel decoder progressively upsamples and fuses feature maps ($\mathbf{f}_{4}$, $\mathbf{c}_{16}$, $\mathbf{c}_{8}$, and $\mathbf{c}_{4}$) to generate per-pixel mask embeddings $\mathbf{F}_{m} {\in} \mathbb{R}^{C_{m}\times H\times W}$, where $C_{m}$ is the embedding dimension.

\subsubsection{Transformer Decoder}
The learnable query features $\mathbf{Q} {\in} \mathbb{R}^{N \times C_{q}}$ are fed into the Transformer decoder, where they are jointly processed with $\mathbf{f}_{4}$ and the prompt embeddings. $N$ and $C_{q}$ denote the number of object queries and the embedding dimension, respectively. Each decoder layer consists of a Prompt-based Restoration Attention, followed by self-attention and a Feed-Forward Network (FFN). Learnable positional embeddings are added to the query features.

\subsubsection{Mask and region class prediction}
After refining the per-pixel embeddings and object queries, $\mathbf{Q}$ is passed through a $3$-layer MLP to obtain per-segment embeddings for mask prediction. The membership score for each pixel belonging to a region is derived by multiplying the per-segment embeddings with per-pixel embeddings $\mathbf{F}_{m}$ and applying a sigmoid function:
\begin{equation}
  \mathbf{M}(x) = \sigma(\mathrm{MLP}(\mathbf{Q})~\mathbf{F}_{m}) \in \mathbb{R}^{N \times H \times W}.
\end{equation}
In parallel, the refined object queries are fed into a linear layer followed by a softmax function to produce posterior class probabilities $\mathbf{P}(x)$ of $K$ classes:
\begin{equation}
  \mathbf{P}(x) = \mathrm{softmax}(\mathbf{Q} \mathbf{W}_{p}^\top) \in [0, 1]^{N \times K},
\end{equation}
where $\mathbf{W}_{p} {\in} \mathbb{R}^{K \times C_q}$ is a learnable weight matrix.

\subsubsection{OoD Scoring}
In contrast to per-pixel classification, the ground truth masks are divided into multiple binary masks, each containing all pixels belonging to a specific class. 
Ultimately, the class probabilities $\mathbf{P}(x)$ are broadcast to all pixels within the region, and the class logits $\mathbf{S}(x)$ are calculated as the product of the mask predictions $\mathbf{M}(x)$ with $\mathbf{P}(x)$:
\begin{align}
    \label{eq:logit_s}
    \mathbf{S}(x) = \sum_{n=1}^N \mathbf{P}_{n}(x) \mathbf{M}_{n}(x).
\end{align}
Due to the unique architecture of mask transformer~\cite{cheng2022masked,ackermann2023maskomaly,nayal2023rba}, the class logits $\mathbf{S}(x)$ aggregate weighted voting scores from all object queries; upon convergence, object queries specialize in predicting a specific class. Thus, the outlier scoring function from \textit{Rejected by All}~\cite{nayal2023rba} is adopted:
\begin{align}
    \label{eq:ood_scoring}
    \mathbf{A}(x) = \mathrm{RbA}(\mathbf{S}(x))= -\sum_{k=1}^K \phi(\mathbf{S}_{k}(x)),
\end{align}
with $\phi$ being the $tanh$ activation function. $\mathbf{A}(x)$ represents the outlier score map, indicating the outliers in the image $\mathbf{x}$.

\subsubsection{Panoramic-oriented Disentanglement}
Existing research indicates that during fine-tuning, a balance must be maintained between mask segmentation and semantic prediction within the mask transformer~\cite{nayal2023rba}. We observe that \emph{PanOoS} faces a similar conflict, and accurate decoding of panoramic per-pixel masks is crucial for segmentation performance under conditions of semantic clarity. Specifically, the precise decoding of foreground pixels (\ie, masks) determines the model's capability to detect outliers. To disentangle the conflict between semantic query and per-pixel mask decoding, we introduce a disentanglement strategy specifically designed for \emph{PanOoS}, and only fine-tune the pixel decoder, the $3$-layer MLP, and the linear layers of POS during the outlier exposure phase.

\subsection{Prompt-based Restoration Attention}
\label{subsec:pra}
Traditional query-based decoding is ineffective at capturing the intricate semantic information of panoramic imagery. This limitation primarily arises from prevalent pixel distortion and background clutter in images, which interfere with the semantic expression of queries, introduce semantic confusion, and ultimately impair the model’s comprehension of image content and segmentation performance. To address this challenge, we propose a Prompt-based Restoration Attention (PRA) module that utilizes the clear semantics embedded in text prompts to guide the semantic learning of queries. A self-adaptive mechanism that dynamically adjusts the learning process based on the semantic information of queries is proposed to enhance the robustness and generalization of semantic extraction.

\begin{figure}[!t]
    \centering
    \includegraphics[width=\linewidth]{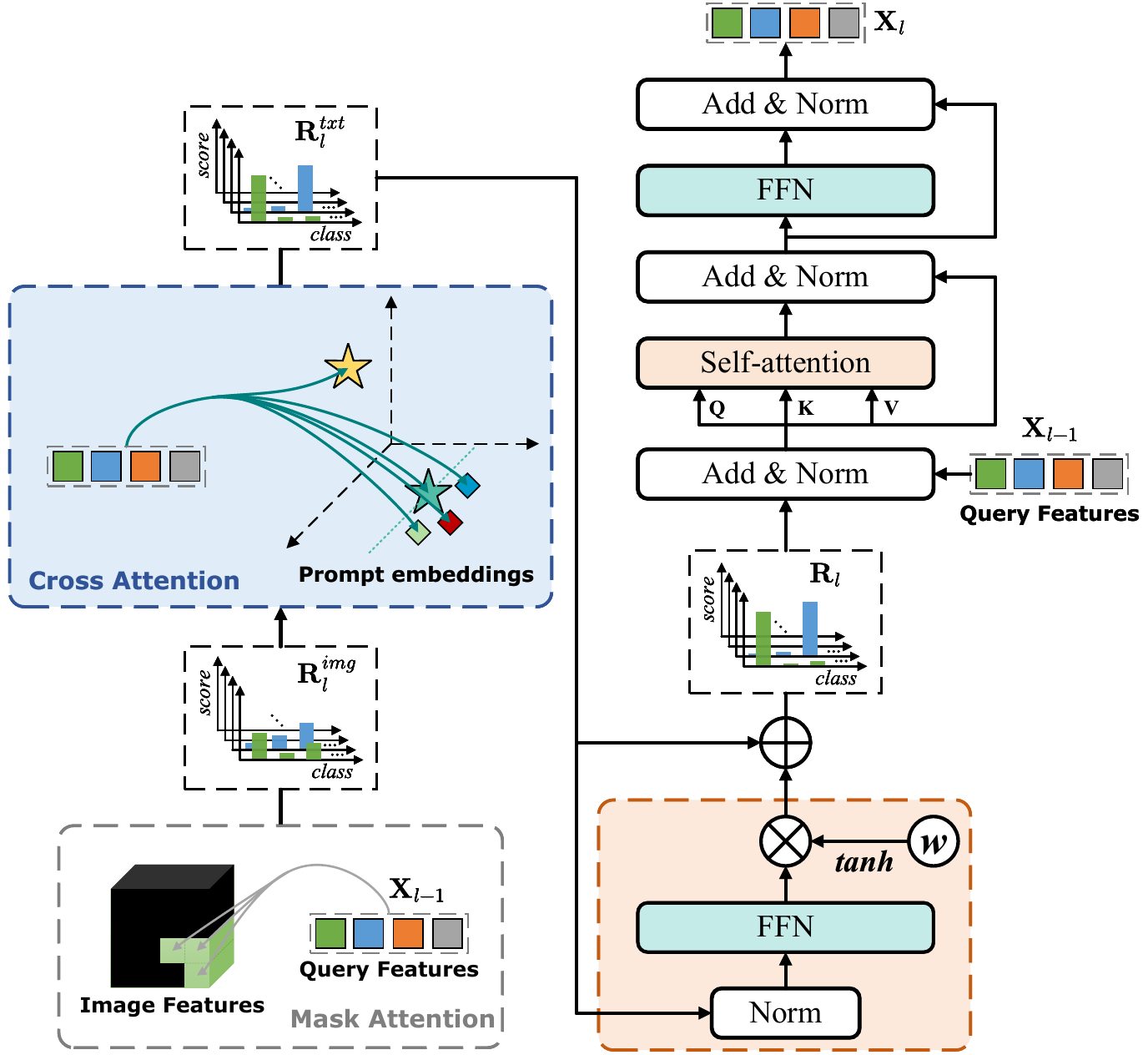}
    \caption{\textbf{Prompt-based Restoration Attention} continuously employs masked attention and cross attention between image features and prompt embeddings while refining semantic clarity through query semantic self-adaptive correction.}
    \label{fig:pra}
    \vskip -2ex
\end{figure}

Specifically, as shown in Fig.~\ref{fig:pra}, PRA refines the query features through two sequential attention operations: masked attention and cross attention. Let $\mathbf{X}_{l-1}\in\mathbb{R}^{N\times C_q}$ denote the query features from the previous decoder layer, where $N$ is the number of queries and $C_q$ is the query dimension. We first update the queries by masked attention with the image features, and then further enhance them by cross-attention with the prompt embeddings. The prompt embeddings include both class-level prompts and distribution-level prompts, denoted as $\mathbf{E}_{t}=[\mathbf{T}_{0},\dots,\mathbf{T}_{K},\mathbf{P}_{in},\mathbf{P}_{out}]\in\mathbb{R}^{(K+3)\times C_m}$, where $\mathbf{T}_{0}$ is the learnable void prompt, $\{\mathbf{T}_{i}\}_{i=1}^{K}$ are the class prompts, and $\mathbf{P}_{in}$ and $\mathbf{P}_{out}$ represent the in-distribution and out-of-distribution prompts, respectively. The two attention operations are formulated as:
\begin{align}
    \mathbf{R}^{img}_{l} 
    &= \mathrm{softmax}(\mathcal{M}_{l-1} + \mathbf{Q}_{l-1}~\mathbf{K}_{i,l}^{T})\mathbf{V}_{i,l}, \\
    \mathbf{R}^{txt}_{l}
    &= \mathrm{softmax}(\mathbf{Q}_{t,l}~\mathbf{K}_{t}^{T})\mathbf{V}_{t},
\end{align}

where $\mathbf{R}^{img}_{l}\in\mathbb{R}^{N\times C_q}$ denotes the image-enhanced queries, and $\mathbf{R}^{txt}_{l}\in\mathbb{R}^{N\times C_q}$ denotes the prompt-enhanced queries. $\mathbf{Q}_{l-1}$ is obtained by linearly projecting $\mathbf{X}_{l-1}$, while $\mathbf{Q}_{t,l}$ is obtained by linearly projecting $\mathbf{R}^{img}_{l}$. $\mathcal{M}_{l-1}$ is the attention mask predicted by the previous decoder layer~\cite{cheng2022masked}. $\mathbf{K}_{i,l},\mathbf{V}_{i,l}\in\mathbb{R}^{H_lW_l\times C_q}$ are the keys and values projected from the image features at the $l$-th decoder layer, where $H_l$ and $W_l$ denote the spatial resolution of the image features. $\mathbf{K}_{t},\mathbf{V}_{t}\in\mathbb{R}^{(K+3)\times C_q}$ are the keys and values projected from the prompt embeddings $\mathbf{E}_{t}$.

Then, we propose a semantic self-adaptive correction mechanism, which dynamically refines the semantic representation of the query, formally expressed as:
\begin{equation}
    \mathbf{R}_{l} =\mathbf{R}^{txt}_{l} + \mathrm{FFN}(\mathrm{LN}(\mathbf{R}^{txt}_{l})) \cdot \tanh(\mathcal{W}).
\end{equation}

Here, $\mathbf{R}_{l}$ denotes the output query features of the $l$-th PRA layer. LayerNorm (LN) is applied to $\mathbf{R}^{txt}_{l}$, and $\mathcal{W}$ represents the learnable \textit{adaptive weight}. We initialize the weight value to zero to ensure the adaptation to diverse multi-scale image features while retaining training stability.

The proposed PRA fully leverages the clearer semantic information in prompt embeddings and incorporates an adaptive semantic correction mechanism to refine the semantic learning process. By mitigating the semantic ambiguity of object queries caused by panoramic pixel distortions and background clutter, PRA substantially improves the accuracy and robustness of query semantic decoding, thereby providing stronger support for panoramic scene understanding.

\subsection{Bilevel Prompt Distribution Learning}
\label{subsec:bpdl}

\begin{figure}[!t]
    \centering
    \includegraphics[width=\linewidth]{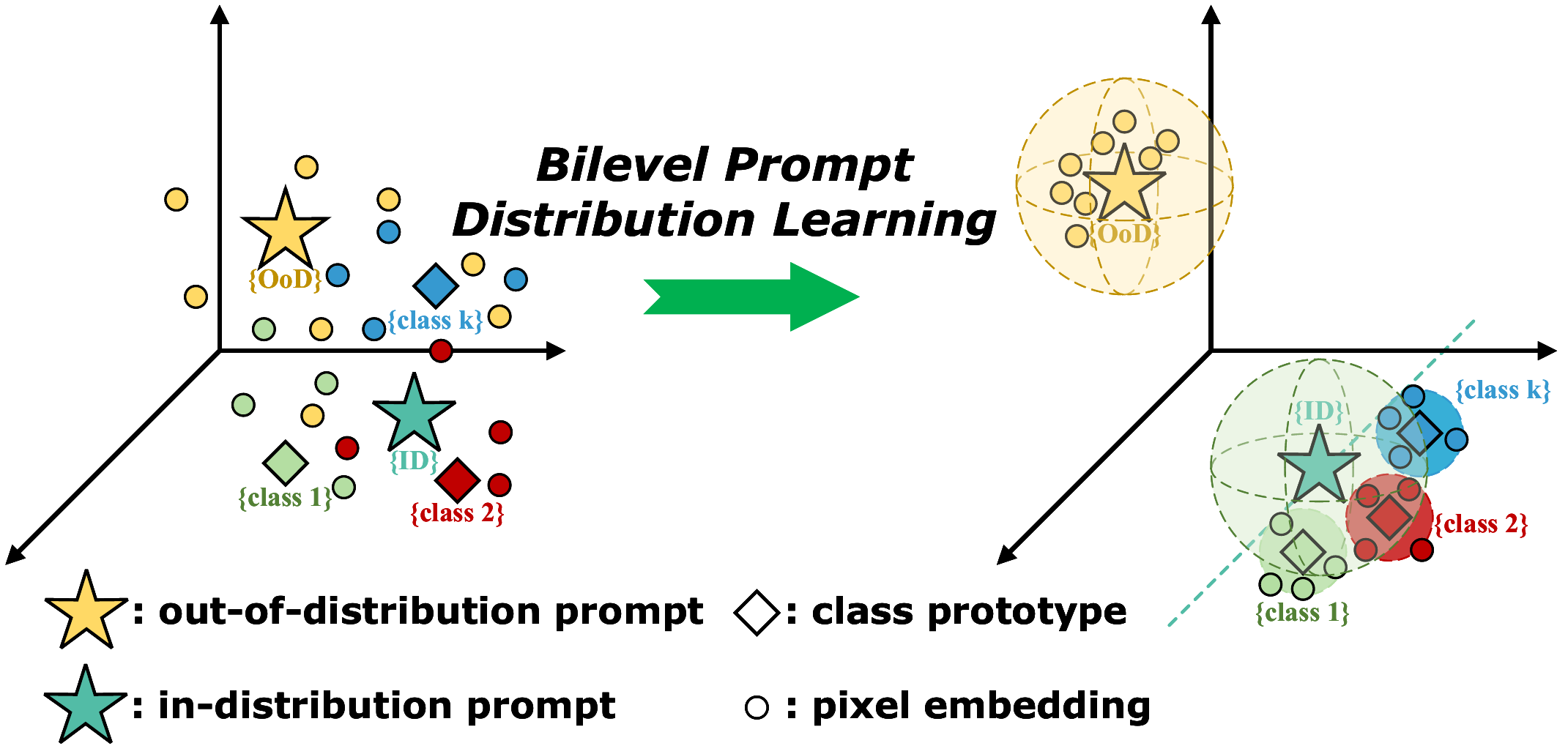}
    \caption{\textbf{Bilevel Prompt Distribution Learning.}
    The distribution prompts, text prompts for each class, and pixel embeddings are denoted by stars, diamonds, and circles, respectively. The dashed circles indicate equidistant lines from the semantic center, and the dashed lines represent the decision boundary separating in- and out-of-distribution regions.}
    \label{fig:bpdl}
    \vskip -2ex
\end{figure}

The pervasive pixel distortions and background clutter in panoramic images often result in dispersed pixel manifolds. This dispersion adversely affects the performance of deep neural networks, which tend to map the unknown samples inside the known class clusters instead of an arbitrary distribution in the whole feature space. Consequently, this misalignment distorts the voting scores of pixel manifolds, particularly in out-of-distribution regions, ultimately leading to overconfident and unreliable predictions. CLIP, trained on large-scale vision-language data pairs, demonstrates robust cross-domain generalizability. Its pretrained text encoder adeptly captures the semantic feature distribution of known classes while maintaining consistency with the pixel space distribution. This accurate semantic feature distribution provides a reliable reference for pixel manifold learning. Leveraging this foundation, we propose a Bilevel Prompt Distribution Learning (BPDL) loss that utilizes projected class and distribution prompts $\{\mathbf{T}_{i} \mid i=0,\dots,K\} \cup \{\mathbf{P}_{in}, \mathbf{P}_{out}\} \in \mathbb{R}^{(K+3)\times C_{m}}$ to optimize the spatial distribution of per-pixel embeddings $\mathbf{F}_{m}$.

As shown in Fig.~\ref{fig:bpdl}, its core objective is to enhance the discriminability of out-of-distribution pixel embeddings $\{\mathbf{O}_{i} \mid i=1,\dots,N_{o}\} \in \mathbb{R}^{N_{o}\times C_{m}}$ near the decision boundary. Specifically, the BPDL optimizes per-pixel embeddings at two levels (in- and out-of-distribution) across four key aspects. For clarity, $N_i$ and $N_o$ denote the numbers of sampled in- and out-of-distribution pixel embeddings, respectively.

\subsubsection{In-distribution learning}
\begin{figure}[!t]
    \centering
    \includegraphics[width=.8\linewidth]{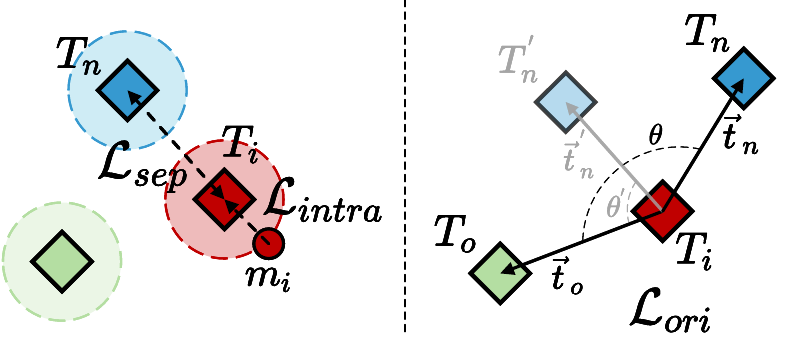}
    \caption{\textbf{In-distribution Learning.} 
    Enhancing intra-class compactness: the similarity of pixel embeddings within the same class is reinforced, resulting in a more compact intra-class distribution. 
    Improving inter-class separation: the embedding distance between different classes is expanded, thereby strengthening the discriminability of inter-class features.}
    \label{fig:inDis}
    \vskip -2ex
\end{figure}

To improve closed-set segmentation performance, particularly by maintaining coherent intra-class representations when distinguishing out-of-distribution from in-distribution pixel embeddings $\{\mathbf{m}_{i} \mid i=1,\dots,N_{i}\} \in \mathbb{R}^{N_{i}\times C_{m}}$, we introduce an in-distribution loss $\mathcal{L}_{pixel}$. 
First, as shown in Fig.~\ref{fig:inDis}, we minimize the Euclidean distance between each pixel embedding and its corresponding class center to enhance the intra-class compactness:
\begin{equation}
    \mathcal{L}_{intra} = \frac{1}{N_{i}}\sum_{i=1}^{N_{i}}\left\Vert \mathbf{m}_{i} - \mathbf{T}_{k_{i}} \right\Vert_2^{2},
\end{equation}
where $k_{i}$ denotes the class label of the embedding $\mathbf{m}_{i}$. Thus, $\mathbf{T}_{k_i}$ is the class-level prototype corresponding to $\mathbf{m}_i$.

As demonstrated in prior works~\cite{he2018triplet,liu2022orientational}, increasing the inter-class distance between category centers helps mitigate semantic ambiguity. Building upon this insight, we propose an inter-class separation loss $\mathcal{L}_{inter}$ to refine the classification boundaries in the feature space. Unlike existing approaches that rely on learned or manually defined class centers, our method leverages text prompts as predefined category prototypes, which inherently exhibit better separation due to their linguistic distinctness. We primarily propose a separation loss $\mathcal{L}_{sep}$ that increases the distance between each class center $\mathbf{T}_{i}$ and its nearest negative class center $\mathbf{T}_{n}$:
\begin{equation}
    \mathcal{L}_{sep} = \frac{1}{K+1}\sum_{i=0}^{K} \max(s - \min_{n\neq i}\left\Vert \mathbf{T}_{i} - \mathbf{T}_{n} \right\Vert_2^{2},\, 0).
    \label{eq:lsep}
\end{equation}
The distance margin $s$ is defined as a hyperparameter.

In addition, to account for the spatial relationships among the remaining negative class centers in the feature space, we introduce an orientation constraint $\mathcal{L}_{ori}$ to optimize the directional relationships of class embedding distributions. 
As illustrated in Fig.~\ref{fig:inDis} (right), when repelling the nearest negative class center $\mathbf{T}_{n}$, the angle $\theta$ between the direction $\vec{t}_{n}$ (from $\mathbf{T}_{i}$ to $\mathbf{T}_{n}$) and the direction $\vec{t}_{o}$ (from $\mathbf{T}_{i}$ to other negative class centers $\mathbf{T}_{o}$) tends to increase:
\begin{equation}
\begin{aligned}
   &\vec{t}_{n} = \mathbf{T}_{n} - \mathbf{T}_{i}, \\
   &\vec{t}_{o} = \mathbf{T}_{o} - \mathbf{T}_{i},
\end{aligned}
\end{equation}
\begin{equation}
   \mathcal{L}_{ori} = \frac{1}{K+1}\sum_{i=0}^{K}\sum_{o\neq i,n} \omega_{o}\left(1 + \cos\left(\vec{t}_{n}, \vec{t}_{o}\right)\right),
\end{equation}
where $\omega_{o}$ represents the weights assigned to other negative class centers $\mathbf{T}_{o}$ (with $o \neq i,n$), which are determined based on their spatial relationships relative to $\mathbf{T}_{n}$:
\begin{equation}
   \omega_{o} = \frac{\exp\left(-\left\Vert \mathbf{T}_n - \mathbf{T}_o \right\Vert_2\right)}{\sum_{k \neq i, n}\exp\left(-\left\Vert \mathbf{T}_n - \mathbf{T}_k \right\Vert_{2}\right)}.
\end{equation}
The inter-class separation loss $\mathcal{L}_{inter}$ is formulated as a combination of two key terms:
\begin{equation}
   \mathcal{L}_{inter} = \mathcal{L}_{sep} + \mathcal{L}_{ori}.
\end{equation}
This acts as a regularization term to further constrain the feature distribution, ensuring structured separation and preventing uncontrolled dispersion of embeddings.

Ultimately, the in-distribution pixel learning loss $\mathcal{L}_{pixel}$ is defined as a sum of the intra-class compactness loss $\mathcal{L}_{intra}$ and inter-class separation loss $\mathcal{L}_{inter}$, formulated as follows:
\begin{equation}
   \mathcal{L}_{pixel} = \mathcal{L}_{intra} + \mathcal{L}_{inter}.
\end{equation}
Minimizing the loss $\mathcal{L}_{pixel}$ enables the model, while capturing the overall feature space distribution, to further refine its ability to differentiate between in-distribution class distributions.

\begin{figure}[!t]
    \centering
    \includegraphics[width=\linewidth]{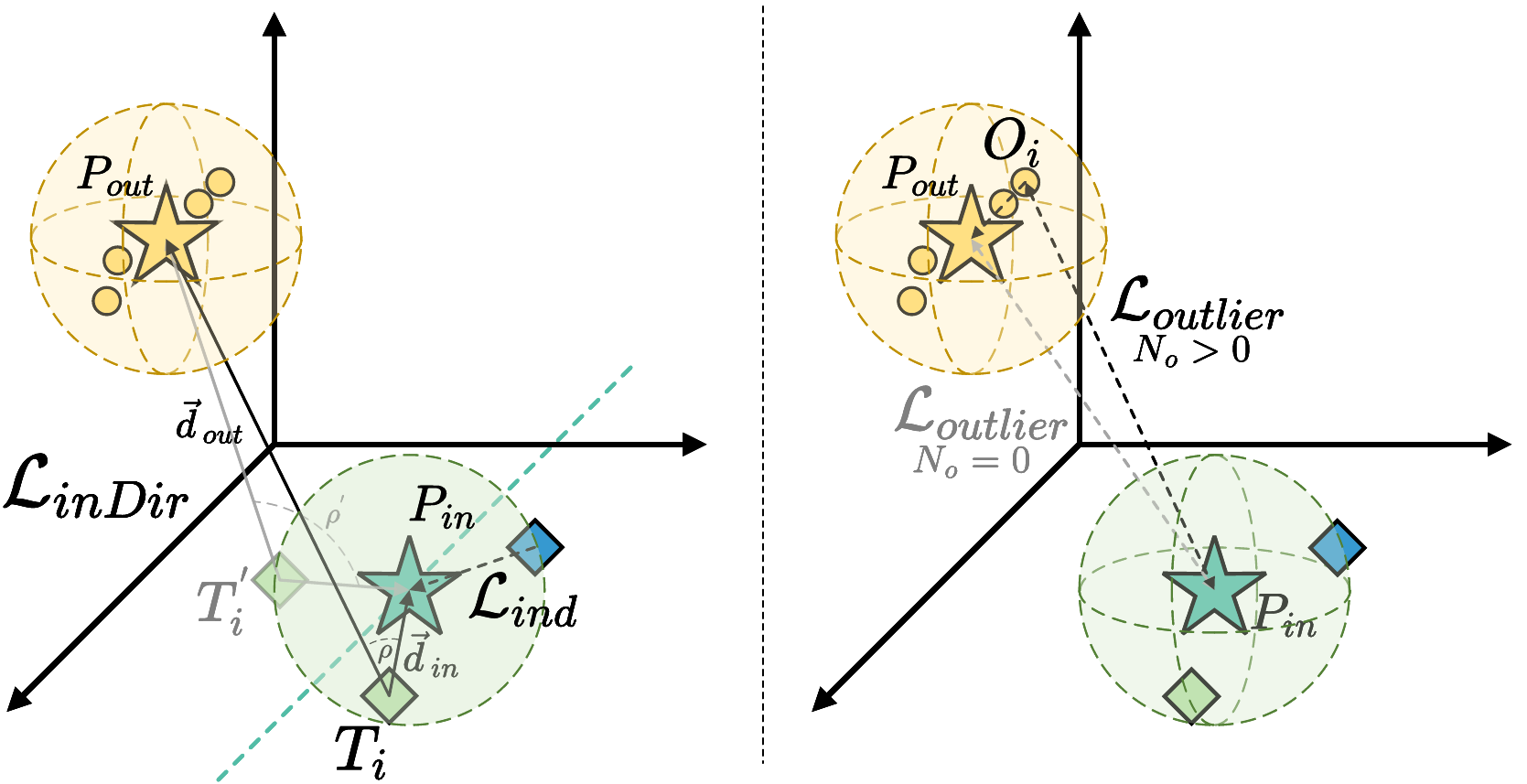}
    \caption{\textbf{Distribution Learning.} 
    Isolation of out-of-distribution pixels: A distinct boundary is established in the feature space to separate out-of-distribution regions from known class regions, thereby mitigating overconfident predictions on outliers. 
    Aggregation of out-of-distribution pixels: Through semantic guidance, out-of-distribution pixel embeddings are clustered into specific semantic spaces, preventing their dispersion and interference with the feature distribution of known classes.}
    \label{fig:outDis}
    \vskip -2ex
\end{figure}

\subsubsection{Distribution learning}
Building upon the aforementioned learning, a higher-level distribution learning mechanism is proposed to enhance the separability of out-of-distribution embeddings $\mathbf{O}_{i}$ and achieve a more precise delineation of the overall feature space. To establish a robust foundation for semantic alignment in subsequent distribution region partitioning and enable the model to more accurately capture the feature distribution patterns of known classes, we first minimize the distance between class text prompts $\{\mathbf{T}_{i} \mid i=0,\dots,K\}$ and in-distribution prompt $\mathbf{P}_{in}$ in the feature space:
\begin{equation}
   \mathcal{L}_{ind} = \frac{1}{K+1}\sum_{i=0}^{K}\left\Vert \mathbf{T}_{i} - \mathbf{P}_{in} \right\Vert_2^{2}.
\end{equation}
The in-distribution compactness loss $\mathcal{L}_{ind}$ promotes alignment between the feature distribution of known classes and the in-distribution semantic information.

To further constrain the distributional discrepancy between known class features and unknown pixel features, we propose a directional constraint strategy. 
As depicted in Fig.~\ref{fig:outDis}, this strategy regularizes the relative directional relationship between each known class prompt and the distribution prompts, reducing the ambiguity between known semantic regions and out-of-distribution regions in the embedding space. 
Specifically, we define the direction from known class embeddings to $\mathbf{P}_{in}$ as $\vec{d}_{in}$ and the direction to $\mathbf{P}_{out}$ as $\vec{d}_{out}$. We optimize these directions to be as aligned as possible by maximizing their cosine similarity. The $\vec{d}_{in}$ and $\vec{d}_{out}$ are formulated as:
\begin{equation}
\begin{aligned}
   \vec{d}_{in} &= \mathbf{P}_{in} - \mathbf{T}_{i}, \\
   \vec{d}_{out} &= \mathbf{P}_{out} - \mathbf{T}_{i}.
\end{aligned}
\end{equation}
The in-distribution directional loss $\mathcal{L}_{inDir}$ is defined as:
\begin{equation}
   \mathcal{L}_{inDir} = \frac{1}{K+1}\sum_{i=0}^{K} \left(1 - \cos\left(\vec{d}_{in}, \vec{d}_{out}\right)\right).
\end{equation}
This approach effectively mitigates confusion between in-distribution and out-of-distribution pixel embeddings in the feature space, thereby enhancing the model's robustness to unknown features. The overall in-distribution loss $\mathcal{L}_{inlier}$ is formulated as the weighted combination of the in-distribution compactness loss $\mathcal{L}_{ind}$ and directional loss $\mathcal{L}_{inDir}$:
\begin{equation}
   \mathcal{L}_{inlier} = \alpha\,\mathcal{L}_{ind} + \mathcal{L}_{inDir},
   \label{eq:linlier}
\end{equation}
where $\alpha$ is the contribution weight hyperparameter of $\mathcal{L}_{ind}$.

Furthermore, to enhance the model's ability to identify out-of-distribution regions, we employ semantic guidance to cluster out-of-distribution pixel embeddings $\{\mathbf{O}_{1},\dots,\mathbf{O}_{N_{o}}\}$ into specific regions of the semantic space. 
During fine-tuning, outliers are pasted into the input images with a preset probability. This outlier exposure strategy simulates the appearance of out-of-distribution samples in real-world scenarios, enabling the model to better adapt to complex tasks in open environments. To facilitate model learning, we propose an out-of-distribution pixel loss $\mathcal{L}_{outlier}$ for both the presence and absence of outliers:
\begin{equation}
\begin{split}
\mathcal{L}_{outlier} &= \left\{
    \begin{aligned}
        &\frac{1}{N_{o}}\sum_{i=1}^{N_{o}} \max(\mathbf{D}_{i,out} - \mathbf{D}_{i,in} + d, ~0), & N_{o}>0\\ 
        &\max(d - \left\Vert \mathbf{P}_{out}-\mathbf{P}_{in} \right\Vert_2^{2}, ~0), & N_{o} =0 
    \end{aligned}
\right.
\end{split},
\label{eq:loutlier}
\end{equation}
where $\{\mathbf{D}_{i,out},~\mathbf{D}_{i,in}\} = \{\|\mathbf{O}_{i}-\mathbf{P}_{out}\|_2^{2},~\|\mathbf{O}_{i}-\mathbf{P}_{in} \|^2_2 \}$, $d$ quantifies the degree of difference between $\mathbf{P}_{in}$ and $\mathbf{P}_{out}$ within the semantic space. 
This approach not only preserves the stability of the known class feature distribution but also provides clear semantic attribution for out-of-distribution features. Additionally, it distinctly delineates out-of-distribution regions in the feature space, establishing a clear boundary with known class regions. As a result, it reduces the model’s overconfidence toward outliers and enhances the reliability of detection. 
The distribution learning loss $\mathcal{L}_{distri}$ is formulated as a sum of the in-distribution loss $\mathcal{L}_{inlier}$ and the out-of-distribution pixel loss $\mathcal{L}_{outlier}$:
\begin{equation}
    \mathcal{L}_{distri} = \mathcal{L}_{inlier} + \mathcal{L}_{outlier}.
\end{equation}

Finally, the BPDL loss $\mathcal{L}_{bpdl}$ is defined as follows:
\begin{equation}
    \mathcal{L}_{bpdl} = \mathcal{L}_{pixel} + \mathcal{L}_{distri}.
\end{equation}
Through the Bilevel Prompt Distribution Learning mechanism, the manifold of per-pixel embeddings for mask predictions becomes more compact and structured, thereby demonstrating better robustness and generalization in \emph{PanOoS}.

\subsection{Training and Inference}
\label{subsec:trainingandInference}
\subsubsection{Training}
Following~\cite{cheng2022masked,nayal2023rba,rai2024mask2anomaly}, during training, bipartite matching is employed to match each ground truth mask with an object query, supervised by region prediction and classification losses, while the fine-tuning phase further incorporates outlier supervision and bilevel prompt distribution learning. 
For region prediction, a weighted combination of binary cross-entropy loss and Dice loss~\cite{milletari2016v}:
\begin{equation}
    \mathcal{L}_{mask}=\lambda_{bce}~\mathcal{L}_{bce} + \lambda_{dice}~\mathcal{L}_{dice},
\end{equation}
is applied to the binary mask predictions, and cross-entropy loss $\mathcal{L}_{cls}$ is used for classification to learn the semantic classes of each mask. 
The total training loss is defined as follows:
\begin{equation}
    \mathcal{L} = \mathcal{L}_{mask} + \lambda_{cls}~\mathcal{L}_{cls}.
\end{equation}
In the outlier exposure fine-tuning phase, we further utilize the $RbA$ loss $\mathcal{L}_{rba}$~\cite{nayal2023rba} to suppress the high-confidence probabilities of outlier pixels in known classes, and the bilevel prompt distribution learning loss $\mathcal{L}_{bpdl}$ to refine the pixel manifold of per-pixel mask embeddings:
\begin{equation}
    \mathcal{L}_{oe} = \mathcal{L} + \mathcal{L}_{rba} + \lambda_{bpdl}~\mathcal{L}_{bpdl},
    \label{eq:loe}
\end{equation}
where $\lambda_{bpdl}$ is the balancing coefficient.

\subsubsection{Inference}
During inference, we do not utilize outlier exposure or BPDL optimization. Given an input panoramic image, the image encoder extracts multi-scale image features, which are fed into the pixel decoder to generate semantically enriched features alongside the per-pixel mask embeddings for segmentation. Concurrently, class-level and distribution-level prompts are embedded through the text encoder and projected into pixel space. These prompt embeddings, along with the decoded image features, are jointly processed by the Transformer decoder with PRA, where they interact iteratively with learned query features. The refined queries are subsequently used for mask prediction and category classification, which are integrated into per-pixel class logits according to Eq.~\ref{eq:logit_s}. Finally, an OoD scoring function in Eq.~\ref{eq:ood_scoring} is applied to produce the out-of-distribution segmentation prediction.

%% file: Tex_content/benchmark.tex
\section{Established Benchmarks}
\label{sec:benchmarks}

\subsection{Overview of the Benchmarks}
\label{subsec:overviewofBench}
In this work, we establish two novel benchmarks: DenseOoS and QuadOoS, specifically designed for \emph{PanOoS}. $360^{\circ}$ panoramas have a broader FoV and many small objects, which exaggerate the costs of creating pixel-wise annotations in unconstrained surroundings. 
From the perspective of the task, we aim to overcome the scarcity of \emph{PanOoS} testbeds and effectively evaluate the impact of mitigating the severe distortions and background clutter on out-of-distribution segmentation performance when unmasking the narrow FoV. Furthermore, to explore \emph{PanOoS} in real-world, unconstrained, unstructured, and dense-motion scenarios, we deploy a quadruped mobile robot to capture images in outdoor motion scenes with outliers, creating a real-world benchmark. Our objective with two benchmarks is to provide a comprehensive evaluation of methods capable of performing \emph{PanOoS}. For comparison with previous pinhole-\emph{OoS} methods, we adopt Area under PRC Curve (AuPRC/AP) and False Positive Rate at a true positive rate of $95\%$ (FPR$_{95}$) for pixel-wise evaluation of \emph{PanOoS}, and mean Intersection-over-Union (mIoU) for closed-set segmentation evaluation. We unfold a comprehensive explanation in Appendix~\ref{sec:evaluationMetrics}.

\input{tables/comparisonDatasets}

\subsection{DenseOoS}
We introduce a novel test dataset: Dense Panoramic Out-of-Distribution Segmentation (DenseOoS) tailored for \emph{PanOoS}. DenseOoS comprises a labeled test set of $1,000$ panoramic images with resolution $2048{\times}400$ and provides pixel-wise annotations for the \emph{PanOoS} task, which greatly extend the semantic labels from DensePASS~\cite{ma2021densepass}. Relevant studies have shown that using generated outlier data for model training or evaluation is an effective approach~\cite{loiseau2025reliability,de2024placing}. Therefore, based on the DensePASS dataset, we use a state-of-the-art image generation technique, POC~\cite{de2024placing}, to place multiple classes of outliers into images while generating annotation labels for these outliers to construct DenseOoS.

\begin{figure}[!t]
    \centering
    \includegraphics[width=\linewidth]{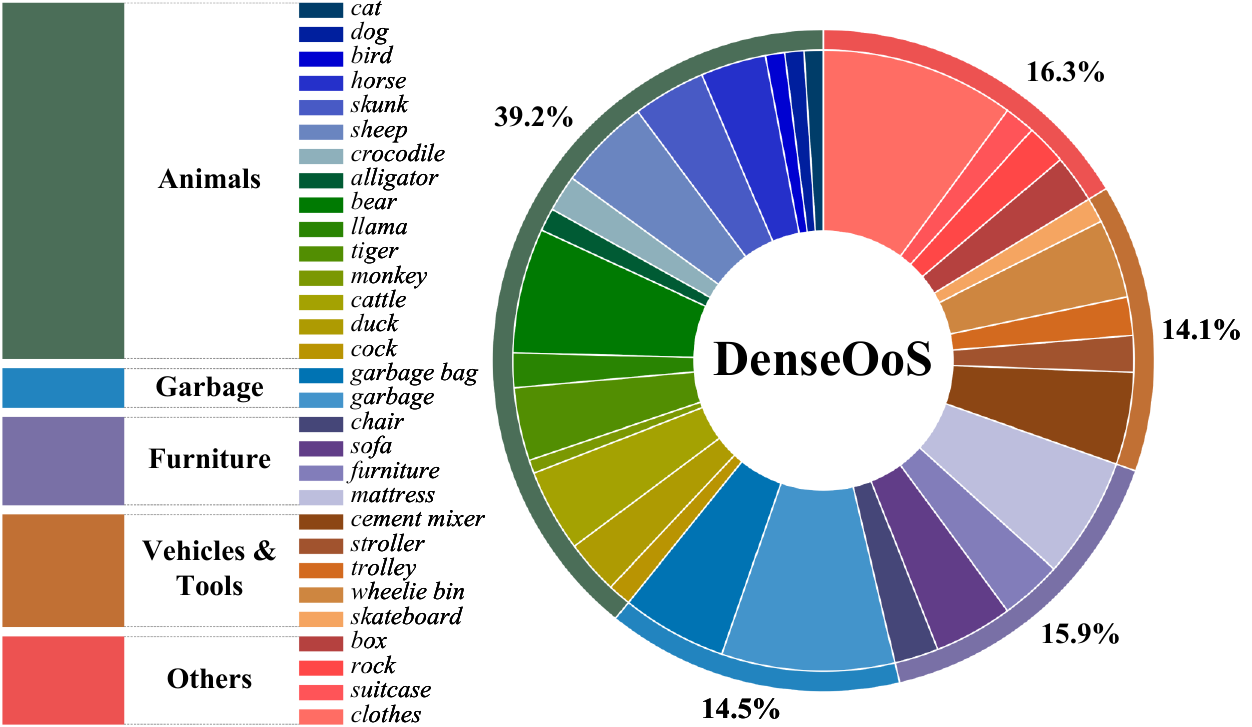}
    \caption{\textbf{Distribution details of DenseOoS.} DenseOoS contains $30$ distinct outliers, divided into $5$ main categories: Animals, Garbage, Furniture, Vehicles~\&~Tools, and Others.}
    \label{fig:denseOoSdetail}
    \vskip -2ex
\end{figure}

The normal labels cover $19$ classes that align with those of Cityscapes~\cite{cordts2016cityscapes}, while the abnormal labels contain $30$ classes of outliers as demonstrated in Fig.~\ref{fig:denseOoSdetail}, such as \emph{animals}, \emph{garbage}, and \emph{furniture}. It additionally includes $5$ other common anomalous objects (\emph{cattle}, \emph{duck}, \emph{cock}, \emph{box}, and \emph{rock}) compared to POC~\cite{de2024placing}. To ensure the high quality and plausibility of the generated images, particularly that the outliers are generated in realistic areas within the real street backgrounds, our screening process follows the ``preliminary screening ${\rightarrow}$ independent screening ${\rightarrow}$ voting'' workflow, eliminating unreasonable object locations, and normal situations. 

\begin{figure*}[!t]
    \centering
    \includegraphics[width=.95\textwidth]{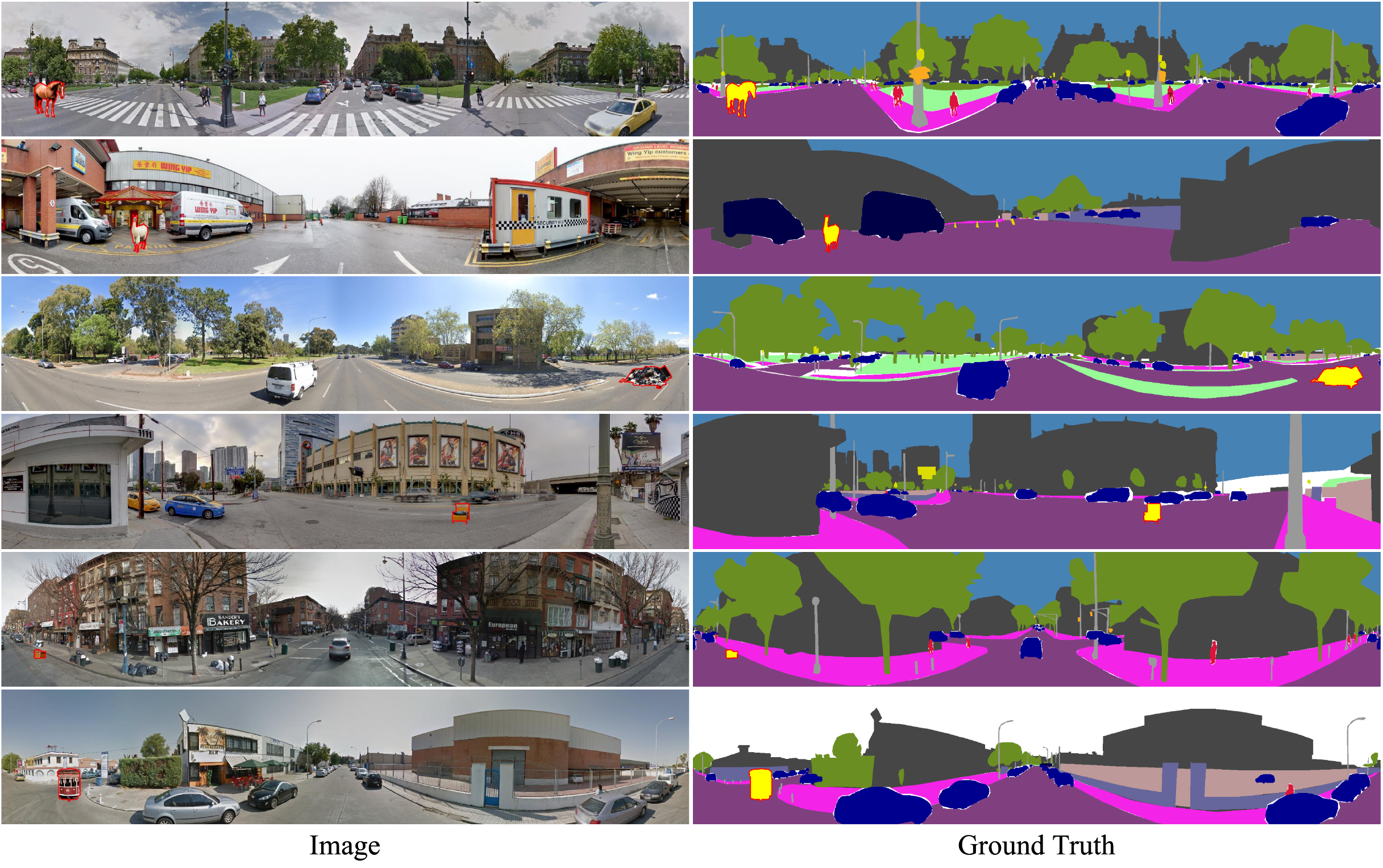}
    \caption{\textbf{Examples from the DenseOoS dataset.} Image backgrounds are highly cluttered, and distortions are severe. Outliers appear in semantically plausible regions. Zoom in for a better view.}
    \label{fig:denseOoSvisual}
    \vskip -2ex
\end{figure*}

The overall process of dataset generation and construction is as follows. First, all text prompts for image inpainting (\emph{a good photo of a $\{class\}$}) are determined based on the DensePASS validation dataset. Next, we select appropriate regions to be inserted (\eg, the \emph{road} and \emph{sidewalk}) and the selected rectangular area is scaled to match the resolution of the real panoramic images. Then, we set different random seeds and use the POC algorithm to batch-generate images and outlier annotations. Finally, the suitable images for constructing the DenseOoS dataset are determined through preliminary manual screening, independent screening, and majority voting. 
Specifically, a total of $14{,}031$ images are generated, of which $1{,}244$ are retained after preliminary manual screening. Following an independent screening and voting process by three examiners, we ultimately select $1{,}000$ images for the construction of the dataset. 
As shown in Fig.~\ref{fig:denseOoSvisual}, the outliers are diverse and distributed within contextually plausible regions of the images.
Importantly, due to the $360^{\circ}$ panoramic layout, over $68\%$ of these outliers lie outside the typical $95^{\circ}$ narrow-FoV of pinhole cameras, which makes DenseOoS particularly suitable for the evaluation of the \emph{PanOoS} task. More details and qualitative examples are provided in the Appendix.

\subsection{QuadOoS}
\label{subsec:quadoos}
\begin{figure*}
    \centering
    \begin{minipage}[b]{0.35\textwidth}
        \subfloat[\small Data collection platform for QuadOoS.]{
            \includegraphics[width=\textwidth]{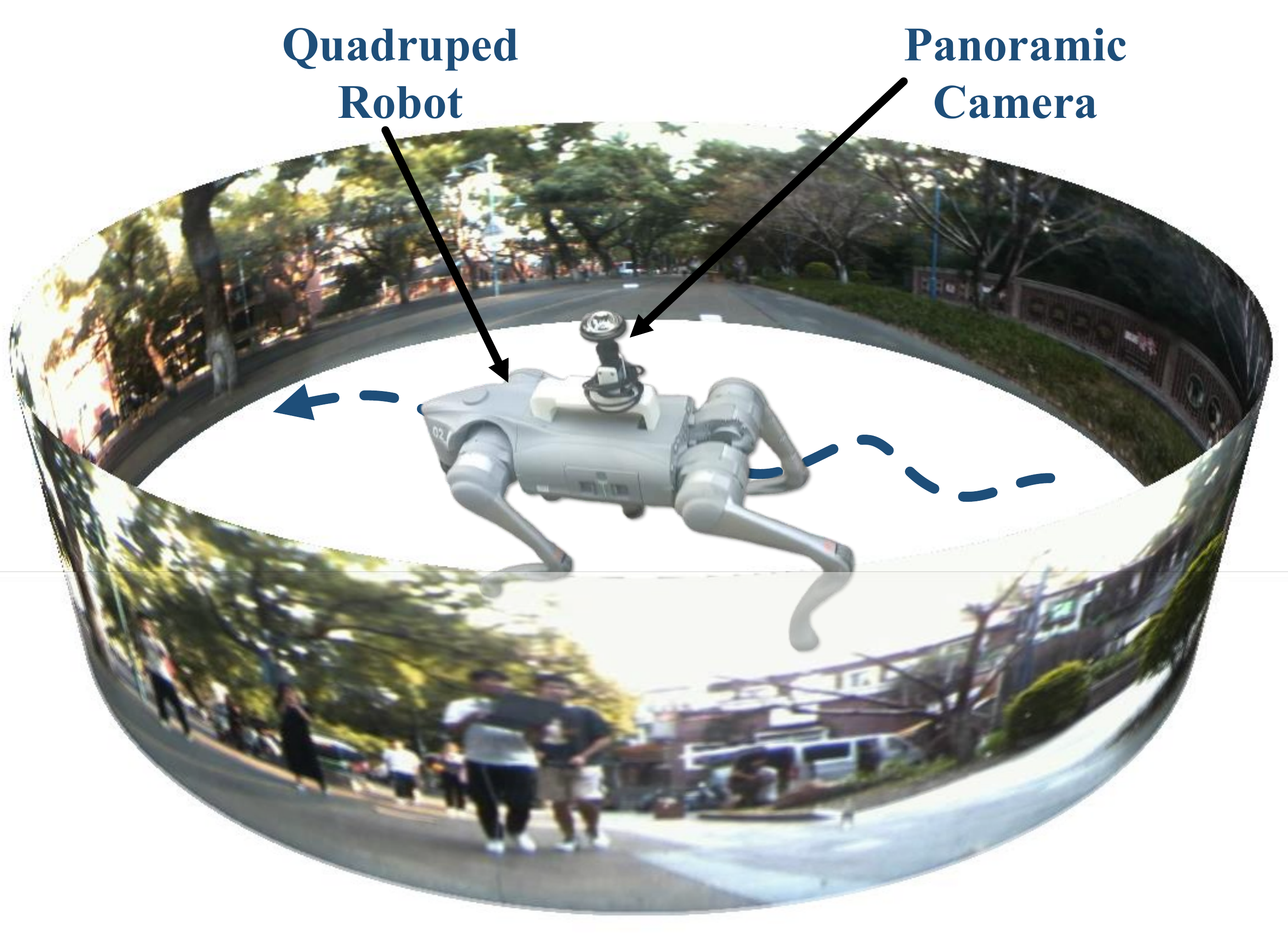}
            \label{fig:robot_dog}}
    \end{minipage}
    \begin{minipage}[b]{0.45\textwidth}
        \centering
        \subfloat[\small Details of the QuadOoS dataset.]{
            \input{tables/quadoosObjectsNum}
            \label{tab:detailsQuadoos}}
        \\
        \subfloat[\small Y-axis pixel positions of motion trajectories.]{
            \includegraphics[width=\textwidth]{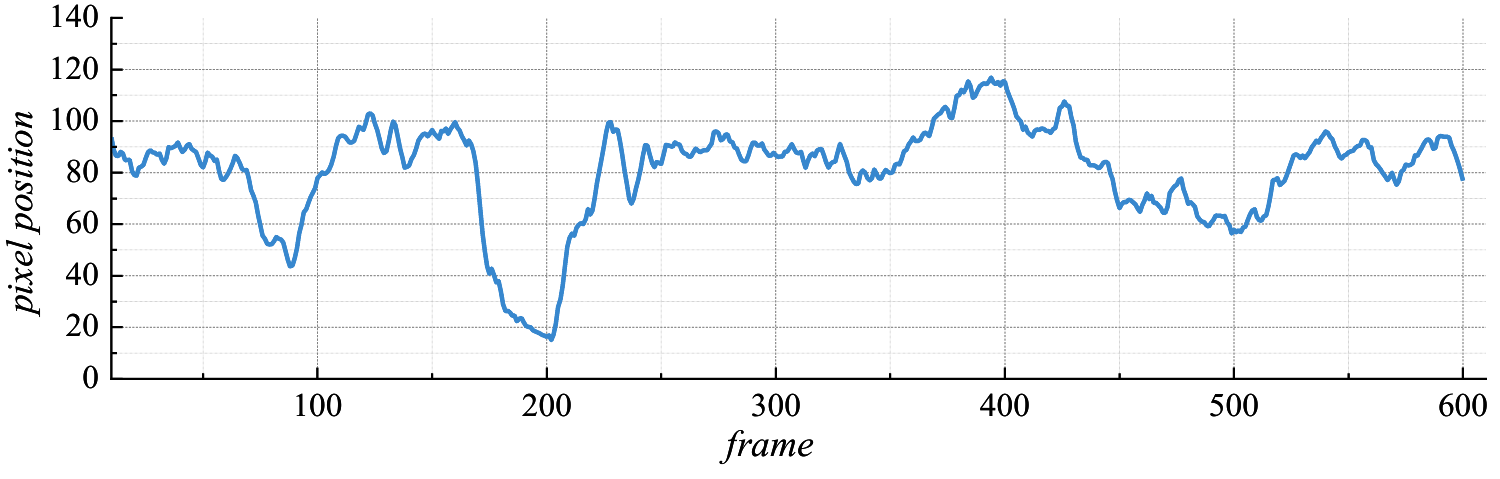}
            \label{fig:ypixel}}
    \end{minipage}
    \caption{(a) Quadruped robot platform with a panoramic annular camera; (b) is the data statistics; (c) Y-axis pixel positions of quadruped robot motion trajectories. While in motion, the quadruped robot experiences significant vertical jitter.}
    \vskip -2ex
\end{figure*}

To further explore the challenges of the \emph{PanOoS} task in real-world applications, we collect panoramic images featuring outliers in unconstrained, unstructured, and dense-motion real scenes, and manually annotate these outliers to establish the first Quadruped Panoramic Out-of-Distribution Segmentation dataset, QuadOoS. This dataset provides a crucial foundation for panoramic out-of-distribution segmentation research and facilitates the evaluation and optimization of \emph{PanOoS} models in complex environments. We develop a mobile platform using a quadruped robot, and the robot is selected for its biomimetic gait, a complex form of locomotion found in nature, which introduces additional challenges for \emph{PanOoS}. It can navigate vertical obstacles up to $15cm$ and inclines up to $30^{\circ}$, making it highly maneuverable in everyday environments. With $12$ joint motors, the robot replicates realistic walking motions at speeds up to $2.5m/s$. For sensing, we use a customized panoramic annular camera to capture a broader FoV ($360^\circ{\times}70^{\circ}$). The camera has a pixel size of $3.45{\mu}m{\times}3.45{\mu}m$, and a resolution of $5$ million effective pixels. Mounted on the quadruped robot (see Fig.~\ref{fig:robot_dog}), the camera ensures an optimal field of view.

Using this platform, we conduct data collection in campus environments with outliers, creating the Quadruped Panoramic Out-of-Distribution Segmentation (QuadOoS) dataset. In particular, we collect data across $8$ blocks in and around the campus, with each block averaging approximately $50m$ in length. The collected raw images consist of a total of $4,800$ frames ($600$ frames per block), each with an original size of $2048{\times}476$ pixels. To avoid misleading the model due to overexposed regions at the center of the lens and to maintain consistency with DenseOoS, we crop the upper portion of the original images (\ie, the \emph{sky}), which does not affect fair evaluation. Subsequently, to ensure the diversity of outliers and the satisfactory imaging quality, we retain a final selection of $106$ original panoramic images after manual screening by five examiners, thereby constructing the QuadOoS dataset. Detailed information is provided in Fig.~\ref{tab:detailsQuadoos}; 
see Appendix~\ref{ap:detailsDatasets} for details of outlier types.

\begin{figure*}[!t]
    \centering
    \includegraphics[width=.95\textwidth]{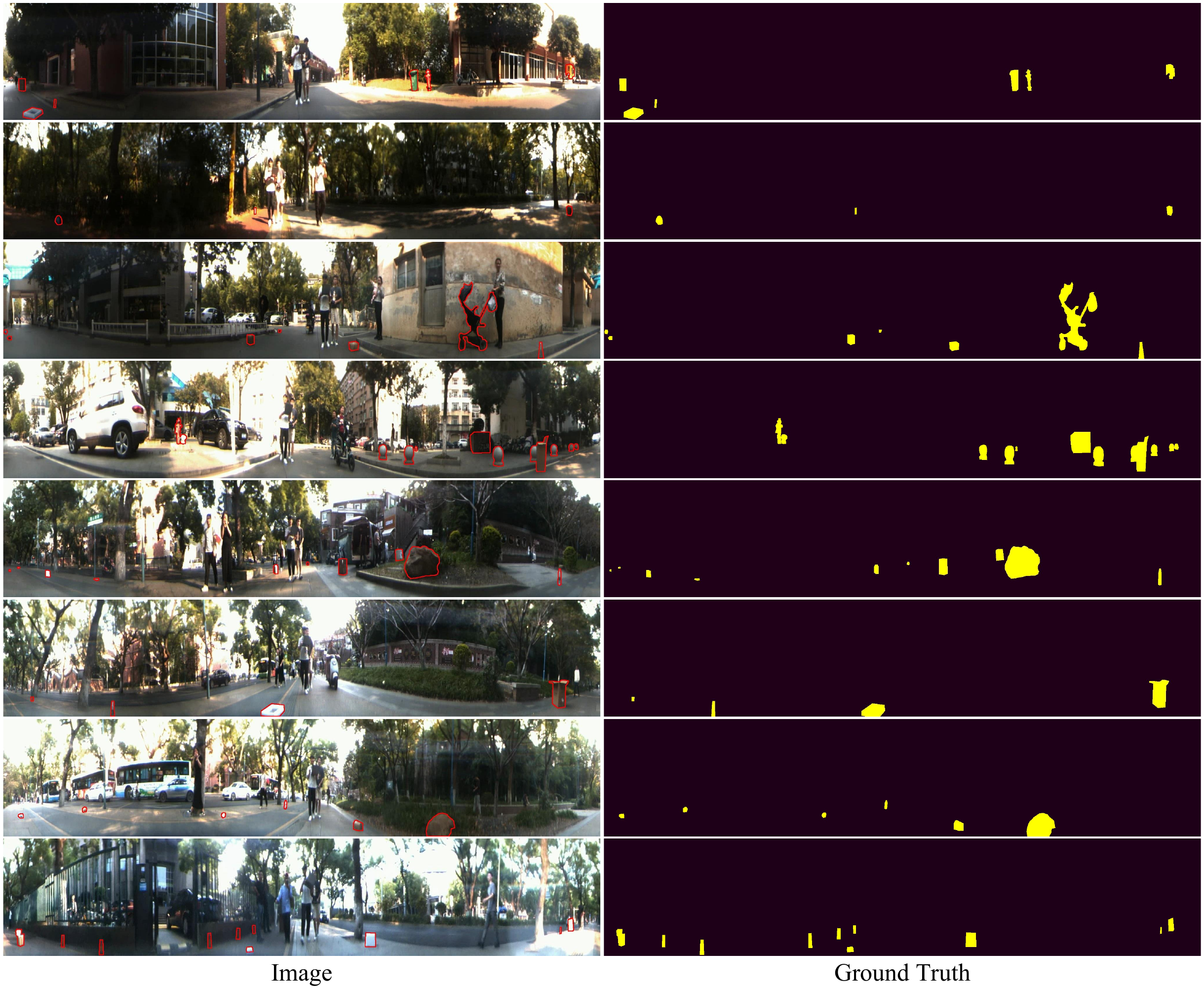}
    \caption{\textbf{Examples from the QuadOoS dataset.}
    The images come with challenges, including uneven exposure, color inconsistencies due to the broader FoV, and increased motion blur. Numerous small outliers are scattered across the scene.}
    \label{fig:quadOoSvisual}
    \vskip -2ex
\end{figure*}

Due to the gait of the quadruped robot, the collected panoramic images exhibit irregular shaking, particularly along the Y-axis (see Fig.~\ref{fig:ypixel}). Compared to panoramic images captured under static conditions, the images in QuadOoS present significant challenges to existing methods, such as uneven exposure, color inconsistencies due to the broader FoV, and increased motion blur, as rapid relative displacement between moving objects and the background intensifies the blurring effect. 
As shown in Fig.~\ref{fig:quadOoSvisual}, each image contains an average of $7$ outliers, accurately reflecting real-world situations and posing substantial challenges for \emph{PanOoS} methods. 
As illustrated in Table~\ref{tab:comparisonDatasets}, with a broader FoV ($360^{\circ}{\times}70^{\circ}$), QuadOoS significantly differs from traditional pinhole-\emph{OoS} datasets~\cite{lis2019detecting,blum2021fishyscapes,hendrycks2019scaling,chan2021segmentmeifyoucan}, and complements DenseOoS by enabling the exploration of motion-induced and panoramic-specific challenges in \emph{PanOoS}.

%% file: tables/comparisonDatasets.tex
\begin{table}[!t]
\centering
\resizebox{\linewidth}{!}{%
\begin{tabular}{@{}lcccccc@{}}
    \toprule
    Datasets & Year  & Domain & Annotation & Test Images & OoD realism & No shift  \\
    \midrule
    Road Anomaly~\cite{lis2019detecting} & 2019 & \multirow{6}{*}{Pinhole} & outlier & 60 & \cmark & \xmark  \\
    Fishyscapes L\&F~\cite{blum2021fishyscapes} & 2019 &  & outlier & 275 & \cmark & \cmarkOrange  \\
    Fishyscapes Static~\cite{blum2021fishyscapes} & 2019 &  & outlier & 1000 & \xmark & \cmark  \\
    StreetHazards~\cite{hendrycks2019scaling} & 2019 &  & all & 1500 & \cmarkOrange & \xmark  \\
    SMIYC RA-21~\cite{chan2021segmentmeifyoucan} & 2021 &  & outlier & 100 & \cmark & \xmark  \\
    SMIYC RO-21~\cite{chan2021segmentmeifyoucan} & 2021 &  & outlier & 327 & \cmark & \xmark  \\
    \hline
    DenseOoS & 2025 & \multirow{2}{*}{\textbf{Panoramic}} & all & 1000 & \cmarkOrange & \cmarkOrange \\
    QuadOoS & 2025 &  & outlier & 106 & \cmark & \cmark \\
    \bottomrule
\end{tabular}
}
\caption{\textbf{Comparison of different OoS datasets.} Referring to POC~\cite{de2024placing}, we qualitatively compare datasets on two main axes, and score them as good (\cmark), medium (\cmarkOrange) or bad (\xmark).}
\label{tab:comparisonDatasets}
\vskip -2ex
\end{table}

%% file: tables/quadoosObjectsNum.tex
\centering
\resizebox{\textwidth}{!}{
\begin{tabular}{@{}lccccc@{}}
    \toprule
    & Frames & Resolution & Locations & Objects & Annotations \\
    \midrule
    Test & 106 & $2048{\times}400$ & 8 & $\ge$18 & 735 \\
    \bottomrule
\end{tabular}}

%% file: Tex_content/experiments.tex
\section{Experiment Results and Analysis}
\label{sec:exper}

\subsection{Datasets}
We train the model on Cityscapes~\cite{cordts2016cityscapes} and an extra inlier dataset: Mapillary Vistas~\cite{neuhold2017mapillary} for outlier exposure fine-tuning. For \emph{PanOoS} evaluation, we use DenseOoS and QuadOoS benchmarks. Additionally, we use Road Anomaly~\cite{lis2019detecting} to evaluate pinhole-\emph{OoS}.
\begin{compactitem}
    \item \textbf{Cityscapes} consists of $2,975$ training and $500$ validation images and contains $19$ categories that are considered inliers in out-of-distribution segmentation benchmarks, such as \emph{road}, \emph{sky}, \emph{building}, and \emph{car}.
    
    \item \textbf{Mapillary Vistas} is a large-scale street-level image dataset containing $25,000$ high-resolution images with a variety of weather, seasons, times of day, cameras, and viewpoints, providing rich visual information.
    
    \item \textbf{Road Anomaly} is a collection of $60$ web images with diverse anomalous objects on or near the road.
\end{compactitem}

\subsection{Implementation Details}
\subsubsection{Closed-Set Training}
We follow \cite{cheng2022masked} and adopt the same training recipe and losses without any special design. The difference is that we adopt the ConvNeXt-Large CLIP~\cite{liu2022convnet} from OpenCLIP~\cite{ilharco_gabriel_2021_5143773}, pre-trained on the LAION-2B dataset~\cite{schuhmann2022laionb}, and freeze its weights. In reference to previous research works~\cite{rai2024mask2anomaly,nayal2023rba}, only one decoder layer with $100$ queries is used in the transformer decoder. Both dimensions $C_{m}$ and $C_{q}$ are set to $256$.
The model is trained for $90k$ iterations using a batch size of $16$ on $4$ NVIDIA GeForce RTX 3090 GPUs. The training is optimized with the AdamW optimizer and a weight decay of $0.05$. We employ the initial learning rate of $1{\times}10^{-4}$ and a polynomial schedule.

\subsubsection{Outlier Exposure Fine-tuning}
After training in the closed-set setting, we fine-tune the pixel decoder, the $3$-layer MLP, and the linear layers for $30k$ iterations on the inlier set of Cityscapes~\cite{cordts2016cityscapes} and Mapillary Vistas~\cite{neuhold2017mapillary}, using the setting of closed-set training; AdamW optimizer with $0.05$ weight decay and $1{\times}10^{-4}$ initial learning rate with polynomial schedule. For each inlier image used in fine-tuning, an object from the MS-COCO samples~\cite{lin2014microsoft} is uniformly chosen as an outlier and pasted on the image using AnomalyMix~\cite{tian2022pixel} with a per-image probability $p_{out}$ (set to $0.3$).

\subsection{Results of Panoramic Out-of-Distribution Segmentation}
We deliver \emph{PanOoS} results on the DenseOoS and QuadOoS benchmarks. Two representative types of existing state-of-the-art pinhole-\emph{OoS} methods, pixel-wise architectures, and mask-transformer-based methods, are benchmarked. As shown in Table~\ref{tab:Panoos}, mask-transformer-based \emph{OoS} methods demonstrate strong potential. 
Compared to the baseline RbA~\cite{nayal2023rba}, POS outstrips it by $34.25\%$ in AuPRC, $21.42\%$ in FPR$_{95}$, and $5.76\%$ in mIoU on DenseOoS. 
Furthermore, state-of-the-art panoramic semantic segmentation models~\cite{hoyer2022daformer,zhang2022bending,zhang2024behind,zheng2024360sfuda++} are compared (see Table~\ref{tab:PSemanticS}). 
Existing panoramic semantic segmentation methods cannot distinguish between in- and out-of-distribution regions and are hindered by outliers, resulting in degraded closed-set segmentation performance. However, POS achieves improved out-of-distribution segmentation while maintaining closed-set segmentation performance, demonstrating the strong potential of prompt distribution learning in omnidirectional scene perception. 
We further explore \emph{PanOoS} in unconstrained, unstructured, and dense-motion real-world scenarios, where all baseline methods struggle to make accurate predictions under motion blur, resulting in significant false positive issues. Specifically, pixel-wise methods are not very sensitive to blur but struggle to effectively recognize outliers within the scene, while mask-transformer-based methods tend to produce false positives. POS achieves a good balance, with mean AuPRC and FPR$_{95}$ improvements of approximately $24\%$ and $13\%$, respectively, on the two benchmarks.
Benefiting from the generalization ability of vision-language models, Prompt-based Restoration Attention (PRA) and Bilevel Prompt Distribution Learning (BPDL), POS performs favorably on the two benchmarks for \emph{PanOoS}, effectively addressing the background clutter and pixel distortions caused by the broader FoV of panoramic images.

\input{tables/panOoS}
\input{tables/panoSemanticSeg}
\input{tables/anomalySegmentation}

\subsection{Results of Pinhole Out-of-distribution Segmentation}
To further investigate the segmentation capacity of the proposed POS in the pinhole domain, we evaluate its OoS performance on one standard pinhole-\emph{OoS} dataset~\cite{lis2019detecting}. 
During the outlier exposure fine-tuning, we do not apply panoramic-oriented disentanglement and fine-tune only the MLP and the linear layers. 
The results in Table~\ref{tab:AnomalySegmentation} indicate that POS excels on Road Anomaly, and achieves precise segmentation of outliers with minimal false positives.

\subsection{Ablation Studies}
\label{subsec:ablationStudy}
In this section, we conduct comprehensive ablation experiments to validate the effectiveness of our solution. To ensure a rigorous comparison, all the results reported in this section are based on the DenseOoS benchmark.

\subsubsection{Each Component in POS}
Table~\ref{tab:components} presents the component-wise ablation results of the technical novelties included in POS. Using the image encoder alone as the backbone, the model demonstrates strong potential for omnidirectional visual perception. However, it still struggles with semantic ambiguity and background interference in complex panoramic scenes. By incorporating Prompt-based Restoration Attention (PRA), the model significantly improves pixel-level semantic associations through prompt distribution learning, enhancing panoramic perception. This results in an increase in AuPRC to $83.94\%$ ($+7.06\%$) and an improvement in mIoU to $62.50\%$ ($+0.77\%$), highlighting the effectiveness in mitigating dense pixel semantic blurring, where prompt-guided attention helps recover clearer query semantics from distorted panoramic image features.

Further incorporation of Bilevel Prompt Distribution Learning (BPDL) leads to a synergistic optimization: PRA enhances query-level semantic restoration, while BPDL preserves topological consistency via manifold learning and provides additional pixel-level distribution regularization for better in-distribution and out-of-distribution separation. 
This combined mechanism enables the model to achieve the best overall performance on DenseOoS: $85.56\%$ AuPRC, $0.45\%$ FPR$_{95}$, and $63.30\%$ mIoU. Not only does it overcome the coupling challenges of semantic ambiguity and geometric distortion that traditional pinhole-\emph{OoS} methods cannot deal with, but it also establishes a new benchmark for panoramic scene understanding.

\input{tables/ablation_Component}
\input{tables/ablation_Pra}

\begin{figure}[!t]
    \centering
    \subfloat[\small MA]{\includegraphics[width=0.32\linewidth]{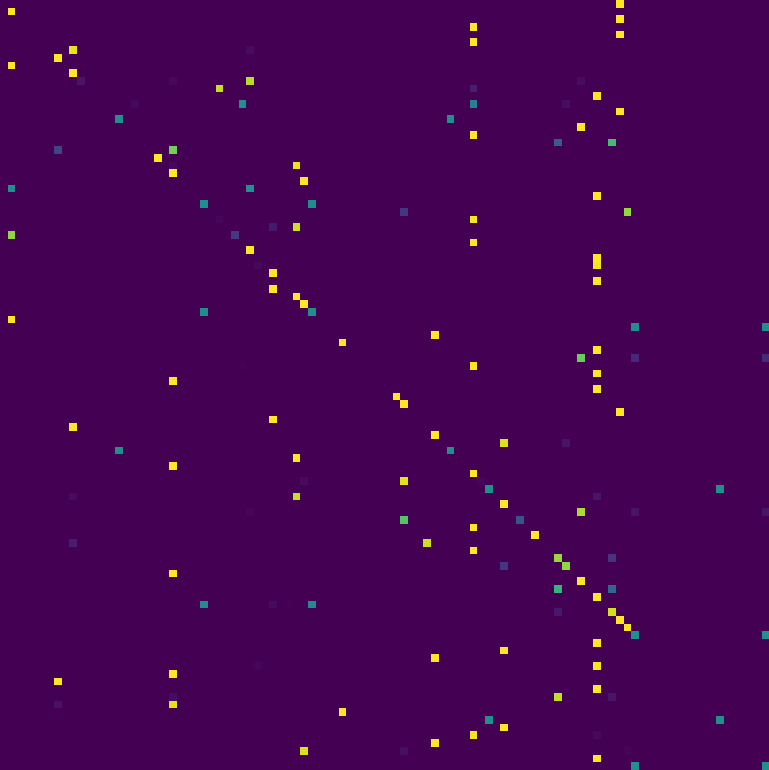}}
    \hspace{1pt}
    \subfloat[\small MA-CA]{\includegraphics[width=0.32\linewidth]{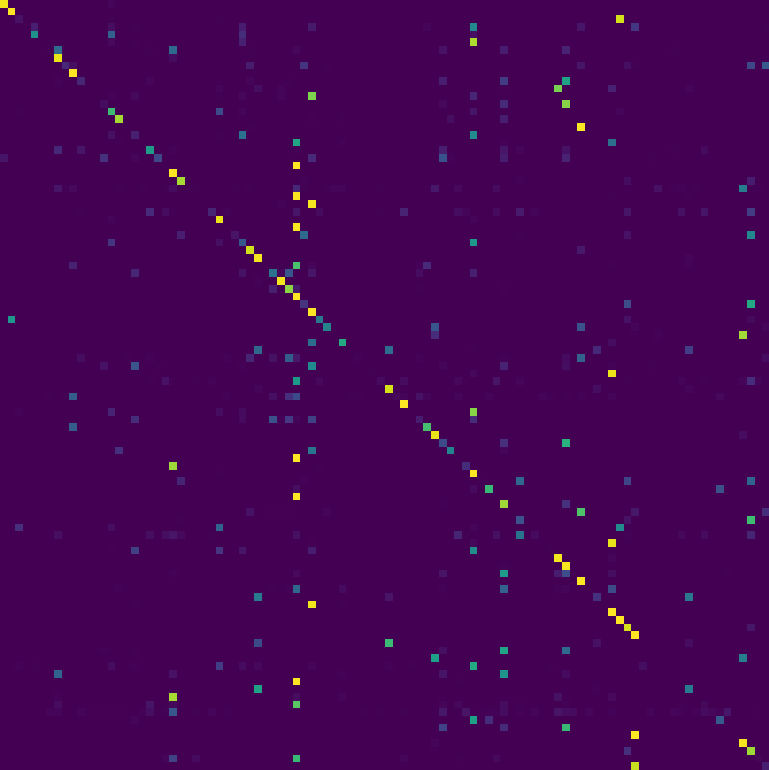}}
    \hspace{1pt}
    \subfloat[\small MA-CA-SAC]{\includegraphics[width=0.32\linewidth]{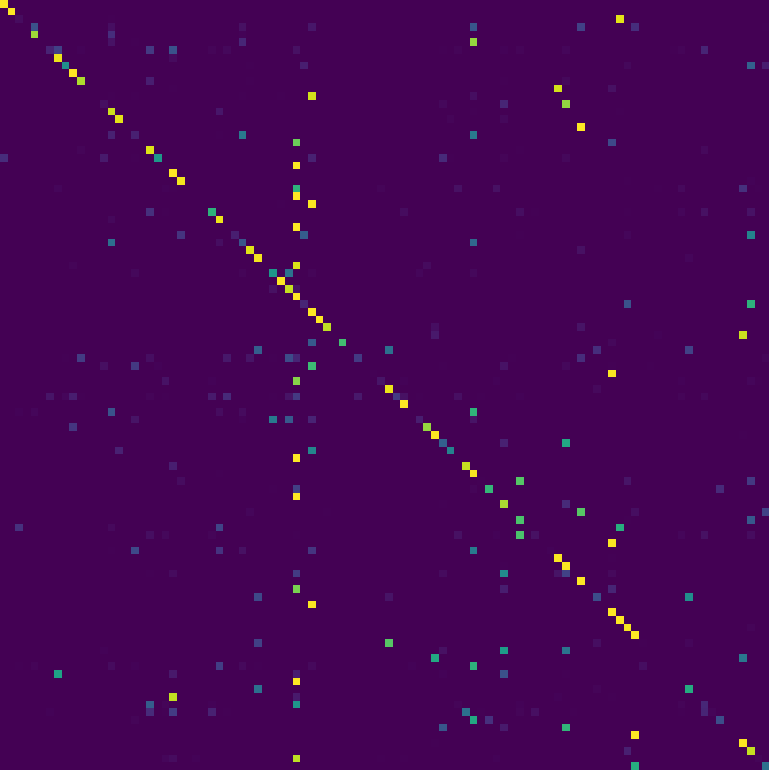}}
    \caption{\textbf{Attention maps of queries} in PRA after different query stages, resembling a confusion matrix in which each query has the highest semantic similarity with itself. When querying only with image features, the semantic information of queries is relatively vague. However, after further interaction with the prompt embeddings, the semantics become clearer, and the self-adaptive correction further mitigates the impact of pixel distortion on semantic decoding.}
    \label{fig:queryMaps}
    \vskip -2ex
\end{figure}

\subsubsection{Prompt-based Restoration Attention}
To better understand the effect of Prompt-based Restoration Attention (PRA), we perform an ablation analysis of its structure in Table~\ref{tab:pra}. Experimental results show that using Masked Attention (MA) alone, which interacts only with image features, cannot adapt to the characteristics of panoramic images, resulting in poor performance across almost all metrics. This demonstrates that prompt-based Cross Attention (CA) effectively captures high-level semantic interactions, alleviating the impact of pixel distortions on semantic decoding. Furthermore, to prevent blind confidence during the learning process, the proposed Self-adaptive Correction (SAC) dynamically adjusts the semantic confidence for each query, yielding more accurate semantic segmentation and avoiding excessively high voting scores for out-of-distribution areas. This is visually substantiated in Fig.~\ref{fig:queryMaps}, where the inter-query attention maps after three stages explicitly manifest the effectiveness of PRA for hierarchical semantic refinement.

\input{tables/ablation_Bpdl}
\begin{figure}[!t]
    \centering
    \subfloat[\small inter-class margin $s$]{\includegraphics[width=0.48\linewidth]{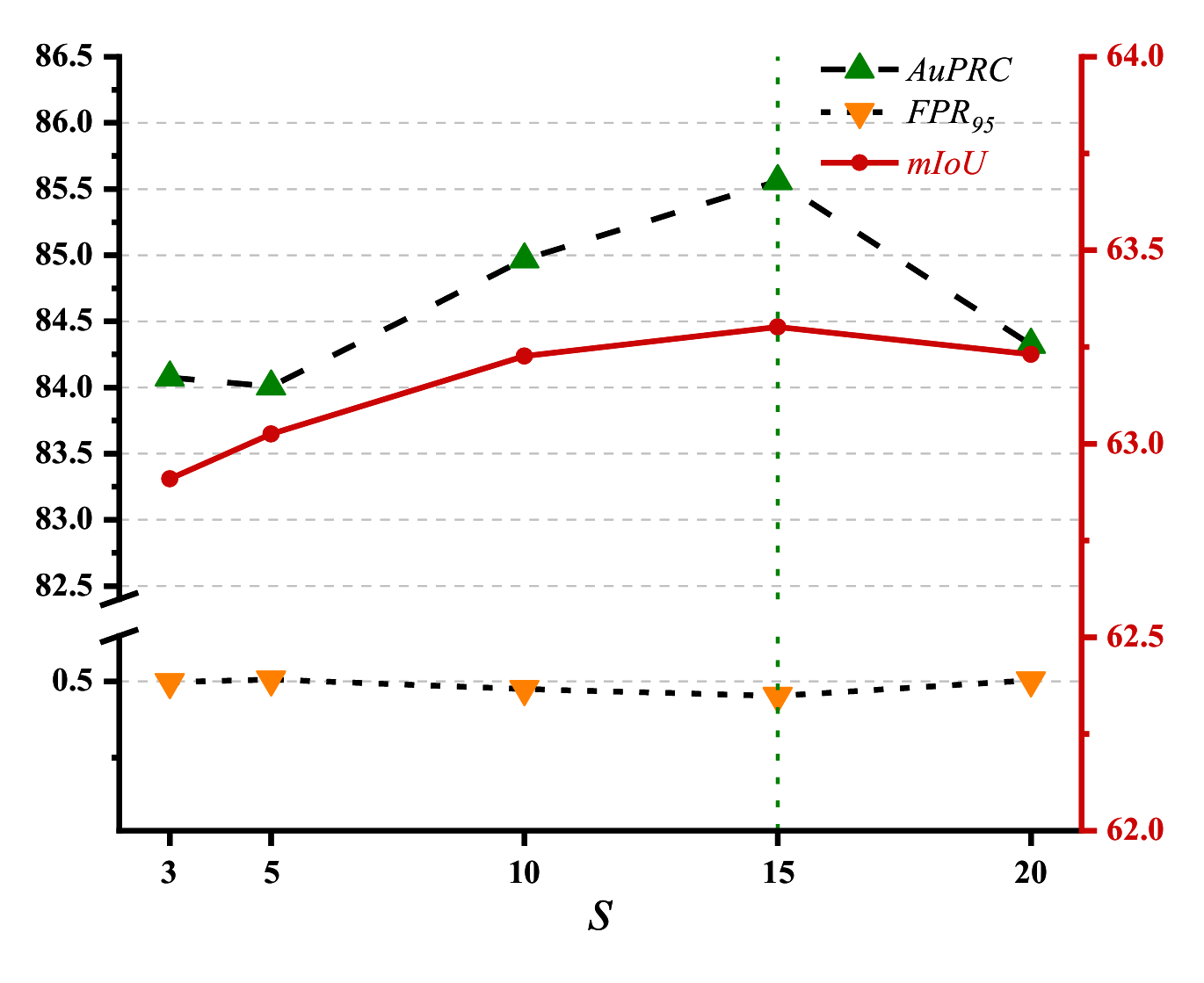}}
    \hspace{1pt}
    \subfloat[\small distribution margin $d$]{\includegraphics[width=0.48\linewidth]{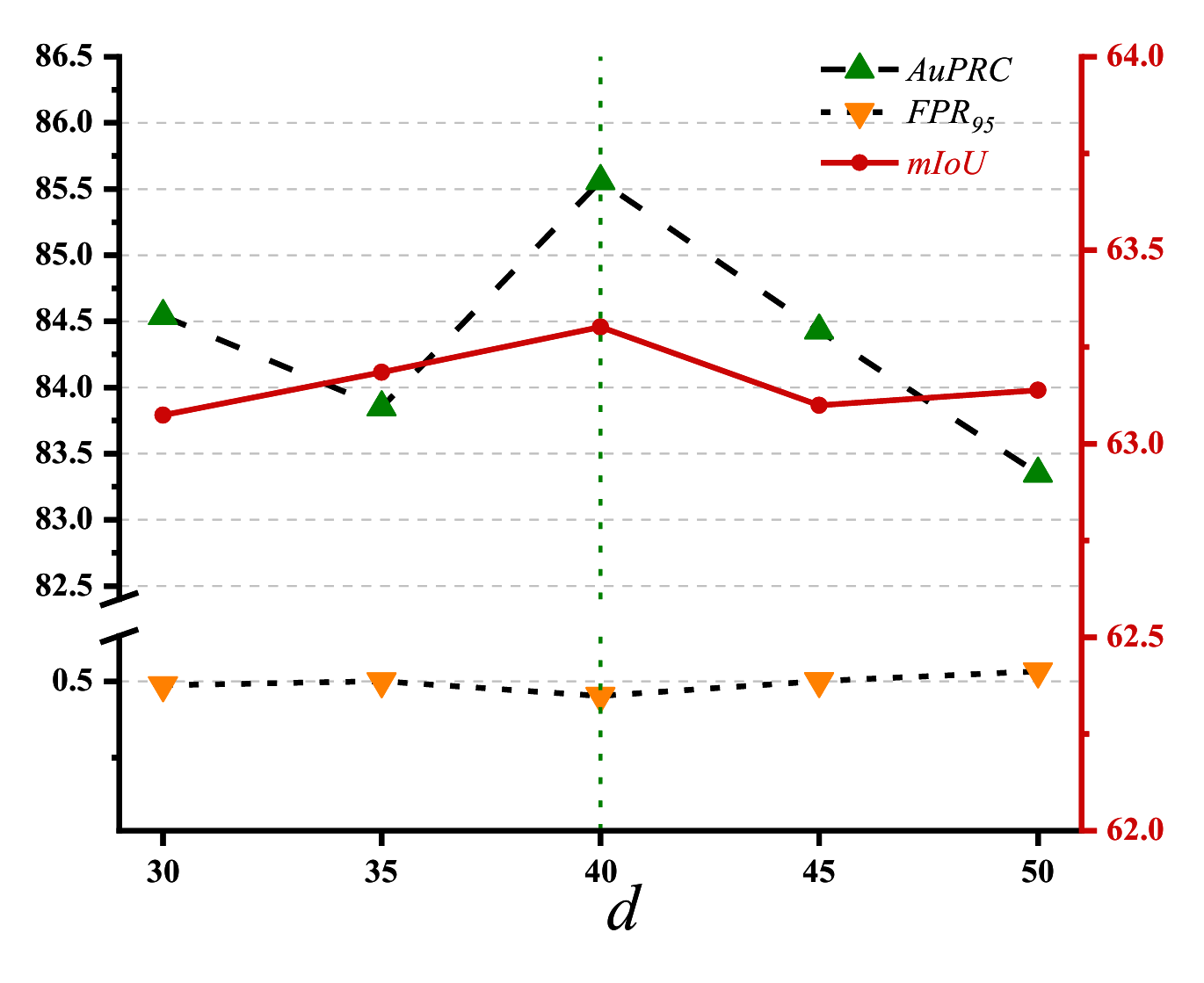}}
    \\
    \subfloat[\small loss weight $\alpha$]{\includegraphics[width=0.48\linewidth]{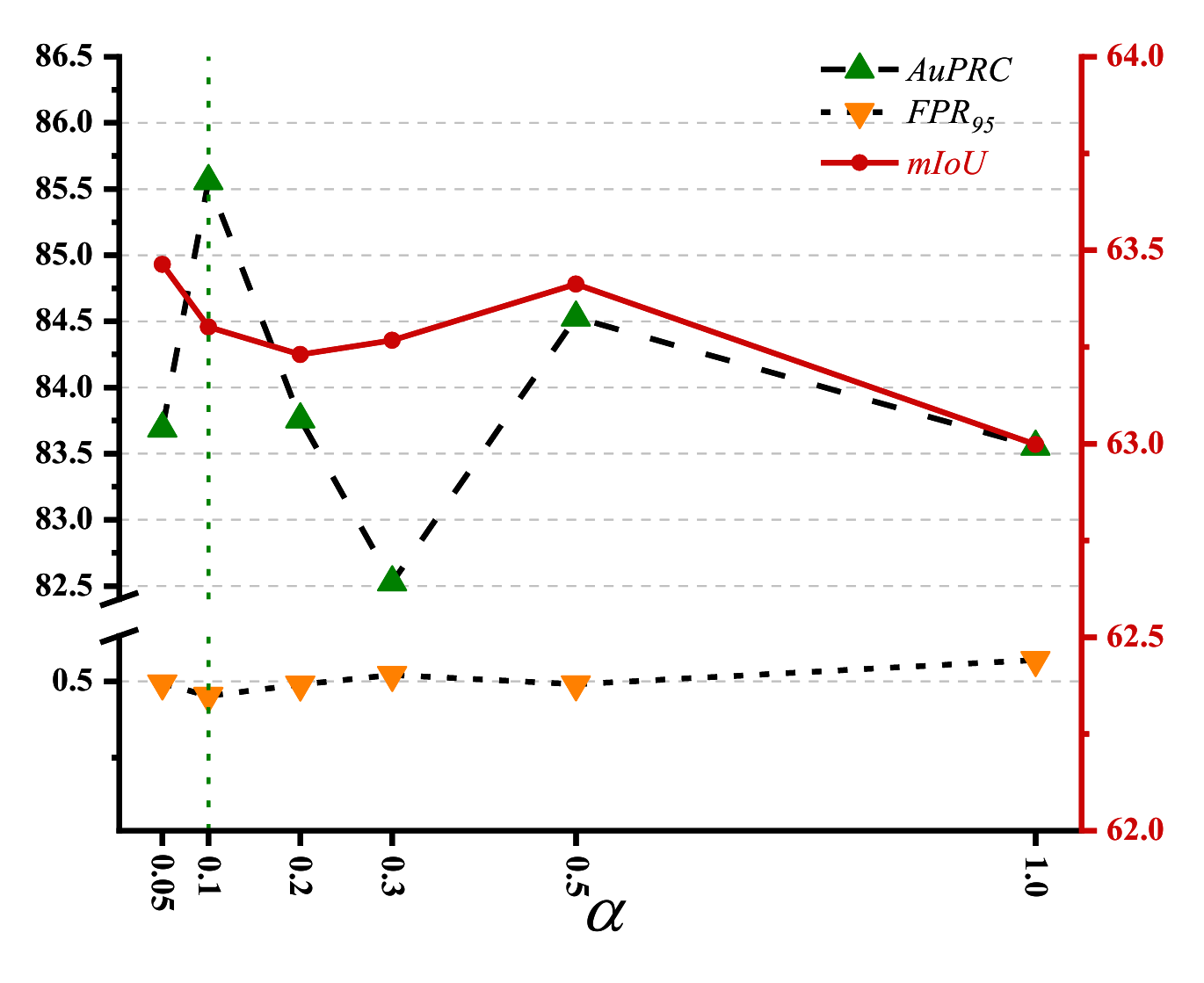}}
    \hspace{1pt}
    \subfloat[\small balancing coefficient $\lambda_{bpdl}$]{\includegraphics[width=0.48\linewidth]{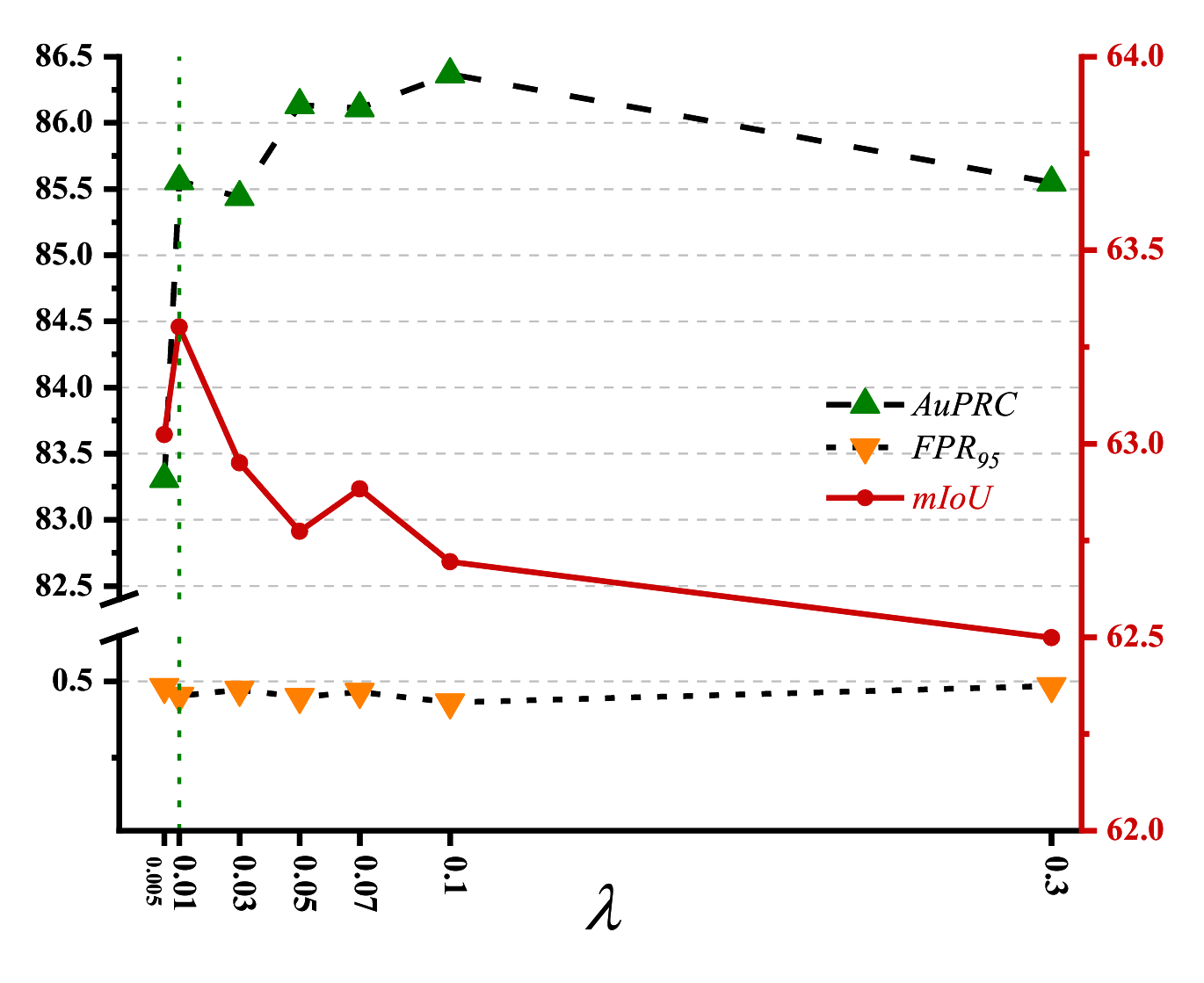}}
    \caption{\textbf{Sensitivity analysis of $s$, $d$, $\alpha$ and $\lambda_{bpdl}$.}}
    \label{fig:hyperpBpdl}
    \vskip -2ex
\end{figure}

\subsubsection{Bilevel Prompt Distribution Learning}
The proposed BPDL belongs to the domain of metric learning. Unlike traditional methods that rely on manually defined or learned centroid embeddings as class prototypes, BPDL uses semantically rich text embeddings as prototypes. This approach allows for a coarse-grained predefinition of distribution regions for each class. Moreover, BPDL utilizes a bilevel learning framework for distribution optimization: the lower-level optimization refines the inter-class distributions for known classes, while the upper-level optimization adjusts the global distribution space. 
As shown in Table~\ref{tab:ablationBpdl}, each component of the $\mathcal{L}_{bpdl}$ contributes to feature distribution optimization, with the $\mathcal{L}_{ind}$ and $\mathcal{L}_{pixel}$ loss terms leading to particularly notable performance gains (AuPRC: ${+}2.95\% / {+}2.48\%$, mIoU: ${+}0.63\% / {+}0.41\%$). Ablation experiments confirm that integrating all loss components results in optimal performance across all evaluation metrics.

As outlined in Eqs.~\ref{eq:lsep}, \ref{eq:linlier}, \ref{eq:loutlier}, and \ref{eq:loe}, the BPDL framework involves four key hyperparameters: inter-class margin $s$, distribution margin $d$, loss weight of $\mathcal{L}_{ind}$ $\alpha$, and balancing coefficient $\lambda_{bpdl}$. The inter-class margin $s$ controls the trade-off between intra-class embedding compactness and inter-class separability, while the distribution margin $d$ governs the geometric structure of the embedding space. The parameter $\alpha$ adjusts the relative contribution of the $\mathcal{L}_{ind}$ loss, and $\lambda_{bpdl}$ balances the weights between the $\mathcal{L}_{bpdl}$ loss and other loss terms. Through a systematic sensitivity analysis (Fig.~\ref{fig:hyperpBpdl}), we observe a notable convex trend for $s$. 
Both mIoU and AuPRC exhibit convex behavior as functions of $s$, while FPR$_{95}$ shows a concave trend but remains stable within a fluctuation range of $0.03\%$. Similarly, the distribution margin $d$ follows a similar optimization trend, confirming the synergistic effect of the bilevel distribution constraints.
The parameter $\alpha$ regulates the distance between the upper-level class prototype embeddings and the distribution embeddings, enhancing the isolation of out-of-distribution embeddings. However, an excessive emphasis on this constraint can reduce closed-set semantic segmentation performance. Additionally, when $\lambda_{bpdl}$ exceeds a certain threshold ($0.01$), the model tends to focus too much on the out-of-distribution detection, leading to significant degradation in closed-set segmentation performance (with mIoU decreasing by $0.6\%$), while out-of-distribution detection metrics remain relatively stable.

\input{tables/ablation_OeFinetuning}

\subsubsection{Outlier Exposure Fine-tuning}
We further empirically evaluate the impact of fine-tuning parameters during the optimization of the prompt distribution space. As shown in Table~\ref{tab:ablationIter}, results on the DenseOoS dataset demonstrate that the model reaches optimal performance after approximately $30k$ iterations, after which overfitting begins. For the outlier data exposure control parameter $p_{out}$, experiments reveal that when $p_{out}{>}0.3$, both closed-set performance and out-of-distribution segmentation performance deteriorate. Based on these results, we selected $0.3$ as the optimal value, balancing outlier detection with closed-set performance.

Moreover, to provide a more comprehensive comparison of the fine-tuning strategies, we additionally report Negative Log-Likelihood (NLL) and Expected Calibration Error (ECE) to assess probabilistic prediction quality and confidence calibration, respectively, together with mIoU on the complete DensePASS~\cite{ma2021densepass} validation set (mIoU$_{\mathrm{DP}}$) to evaluate ID semantic segmentation on clean panoramic images. 
As shown in Table~\ref{tab:ablationStr}, the panoramic-oriented disentanglement strategy, denoted as Pixel Dec., achieves the best overall performance across the six evaluation metrics. 
Specifically, fine-tuning only the MLP and linear layers (\emph{w/o}) or additionally fine-tuning the Transformer decoder (Trans. Dec.) fails to effectively address the complexities of panoramic images. 
Although jointly fine-tuning all decoder components yields a comparable FPR$_{95}$, its broader optimization scope increases the number of trainable parameters during OE fine-tuning from $7.07$M to $9.49$M, while providing no improvement across the performance metrics and yielding lower semantic segmentation accuracy.
In contrast, fine-tuning the pixel decoder, the 3-layer MLP, and the linear layers achieves the most favorable overall performance across OoD detection, semantic segmentation, probabilistic prediction quality, and confidence calibration. These findings verify the effectiveness of the panoramic-oriented disentanglement strategy in enhancing out-of-distribution detection while maintaining closed-set segmentation performance.

\begin{figure}
    \centering
    \includegraphics[width=\linewidth]{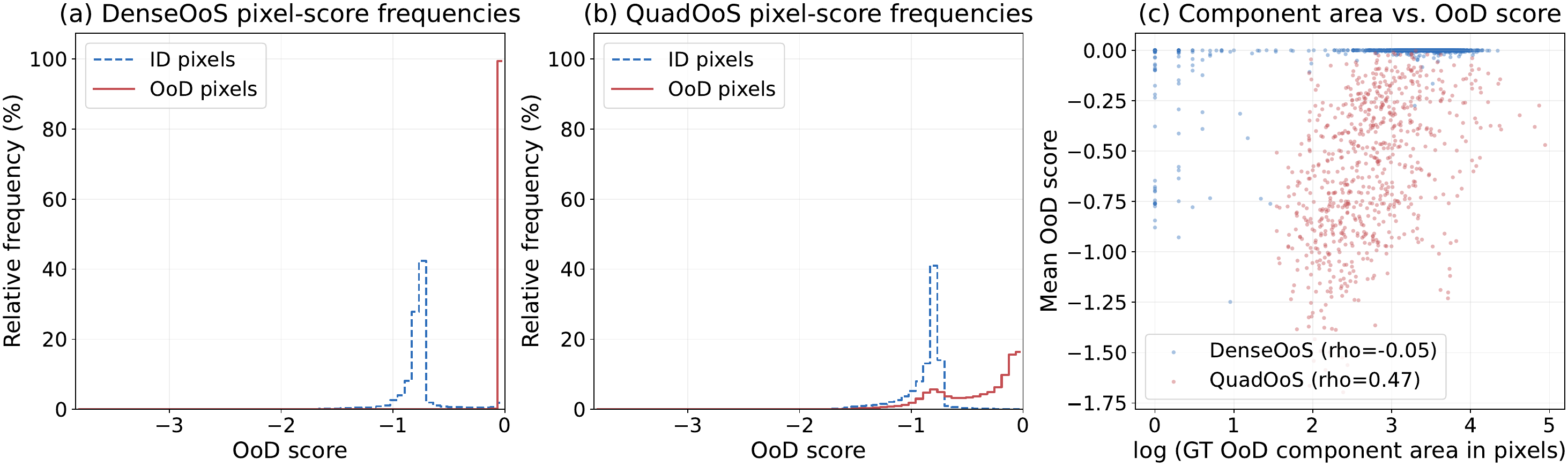}
    \caption{\textbf{OoD score statistics on DenseOoS and QuadOoS.} (a) and (b) show the relative-frequency distributions of ID and OoD pixel scores. (c) shows the relationship between the logarithm of GT OoD-component area and mean OoD score, together with the Spearman correlation $\rho$.
    }
    \label{fig:oodScore}
    \vskip -2ex
\end{figure}

\subsection{Out-of-Distribution Score Analysis}
\label{appendix:score_statistics}
To characterize the difference between outliers in DenseOoS and QuadOoS, we analyze the raw OoD scores produced by POS. Invalid pixels are excluded, while no dataset-level score normalization or pixel subsampling is applied. For visualization, the in-distribution (ID) and OoD histograms are separately normalized by their respective pixel counts. According to the scoring function in Eq.~\ref{eq:ood_scoring}, scores closer to zero indicate stronger OoD evidence.

As shown in Fig.~\ref{fig:oodScore}a, the OoD pixels in DenseOoS are concentrated near the maximum OoD response and are clearly separated from the ID distribution. By contrast, Fig.~\ref{fig:oodScore}b shows that real-world OoD pixels in QuadOoS have a broader score distribution and overlap more substantially with ID pixels. Nevertheless, the mean score of OoD pixels on QuadOoS remains higher than that of ID pixels ($-0.404$ \vs~$-0.876$), indicating that POS retains meaningful discriminative capability on real-world panoramic data.
We further analyze the relationship between the area of each ground-truth OoD connected component and its mean OoD score. As shown in Fig.~\ref{fig:oodScore}c, DenseOoS exhibits almost no monotonic relationship between component area and mean OoD score, with a Spearman correlation of $\rho{=}-0.05$. In contrast, QuadOoS exhibits a moderate positive correlation of $\rho{=}0.47$, indicating that larger real-world outliers tend to receive stronger OoD responses, whereas extremely small outliers are more likely to be confused with surrounding ID regions.

However, the observed difference should not be attributed solely to the synthetic appearance of outliers. Compared with DenseOoS, QuadOoS additionally introduces a different panoramic imaging system, uneven exposure, motion blur, complex backgrounds, and a substantially larger proportion of small outliers. The comparison therefore reflects a combined synthetic-to-real appearance gap and an acquisition-domain gap. DenseOoS and QuadOoS serve complementary purposes: DenseOoS provides a scalable and controlled benchmark with dense annotations, whereas QuadOoS serves as a real-world stress test under unconstrained imaging conditions.

%% file: tables/panOoS.tex
\begin{table*}[!t]
    \centering
    \begin{tabular}{@{}lccc|cc|cc@{}}
        \toprule
        \multirow{2}{*}{Method} & \multicolumn{3}{c}{DenseOoS} & \multicolumn{2}{c}{QuadOoS} & \multicolumn{2}{c}{Mean} \\
        & AuPRC $\uparrow$ & FPR$_{95}$ $\downarrow$ & mIoU $\uparrow$ & AuPRC $\uparrow$ & FPR$_{95}$ $\downarrow$ & AuPRC $\uparrow$ & FPR$_{95}$ $\downarrow$ \\
        \midrule
        PEBAL~\cite{tian2022pixel} & 16.39& 7.59& 46.29 & 11.34 & \underline{39.25} & 13.87& \underline{23.42}\\
        DenseHybrid~\cite{grcic2022densehybrid} & 24.44& 7.74& 45.19 & 19.09 & \textbf{33.86} & 21.77&\textbf{20.80}\\ \hline
        EAM~\cite{grcic2023advantages} & 17.97& \underline{6.46}& \underline{61.58} & 33.05& 79.66& 25.51& 43.06\\
        Mask2Anomaly~\cite{rai2024mask2anomaly} & 36.28 & 8.79 & 48.10 & 25.93& 92.55& 31.10 & 50.67\\
        RbA~\cite{nayal2023rba} & \underline{51.31}& 21.87& 57.54& \underline{34.12}& 90.34& \underline{42.71}& 56.11\\
        Ours & \textbf{85.56}& \textbf{0.45}& \textbf{63.30}& \textbf{48.13}& 85.22& \textbf{66.85}& 42.84\\
        \bottomrule
    \end{tabular}
    \caption{\textbf{Panoramic out-of-distribution segmentation quantitative evaluation.} POS shows significant improvement over baseline per-pixel and mask-transformer-based pinhole-\emph{OoS} methods on \emph{PanOoS}.
    Higher values for AuPRC and mIoU are better, whereas for FPR$_{95}$ lower values are better.
    The best and second best results are \textbf{bold} and \underline{underlined}, respectively, and the line in-between the table divides the per-pixel architectures and mask-transformer-based methods.}
    \label{tab:Panoos}
    \vskip -2ex
\end{table*}

%% file: tables/panoSemanticSeg.tex
\begin{table}
    \centering
    \begin{tabular}{@{}lc@{}}
        \toprule
        Method & mIoU \\
        \midrule
        DAFormer~\cite{hoyer2022daformer} & 44.53 \\
        Trans4PASS (Tiny)~\cite{zhang2022bending}  & 44.93 \\
        Trans4PASS (Small)~\cite{zhang2022bending} & 49.35 \\
        Trans4PASS+ (Tiny)~\cite{zhang2024behind}  & 49.29 \\
        Trans4PASS+ (Small)~\cite{zhang2024behind} & 50.11 \\
        360SFUDA++~\cite{zheng2024360sfuda++} & \underline{50.46} \\
        Ours & \textbf{63.30}\\
        \bottomrule
    \end{tabular}
    \caption{\textbf{Quantitative closed-set segmentation results} on DenseOoS. POS achieves the highest mIoU among the compared methods.}
    \label{tab:PSemanticS}
\end{table}

%% file: tables/anomalySegmentation.tex
\begin{table}
    \centering
    \begin{tabular}{@{}lcc@{}}
        \toprule
        Method & AuPRC &  FPR$_{95}$ \\
        \midrule
        PEBAL~\cite{tian2022pixel} & 62.37 & 28.29 \\
        DenseHybrid~\cite{grcic2022densehybrid} & - & - \\
        \hline
        EAM~\cite{grcic2023advantages} & 69.40 & 7.70 \\
        Mask2Anomaly~\cite{rai2024mask2anomaly} & 79.70 & 13.45 \\
        RbA~\cite{nayal2023rba} & \underline{85.42} & \underline{6.92} \\
        Ours & \textbf{89.39}& \textbf{6.25}\\
        \bottomrule
    \end{tabular}
    \caption{\textbf{Pinhole out-of-distribution segmentation results} on Road Anomaly. The best and second best results are \textbf{bold} and \underline{underlined}, respectively; ``-'' indicates the unavailability of benchmark results.}
    \label{tab:AnomalySegmentation}
    \vskip -2ex
\end{table}

%% file: tables/ablation_Component.tex
\begin{table}[!t]
    \centering
    \begin{tabular}{@{}ccccc@{}}
        \toprule
        PRA& BPDL& AuPRC& FPR$_{95}$& mIoU\\
        \midrule
        & & 76.88& 0.84& 61.73\\
        \ding{51}& & 83.94& 0.46& 62.50\\
        \ding{51}& \ding{51}& \textbf{85.56}& \textbf{0.45}& \textbf{63.30}\\
        \bottomrule
    \end{tabular}
    \caption{\textbf{Component-wise ablation study of POS.} PRA: Prompt-based Restoration Attention; BPDL: Bilevel Prompt Distribution Learning.}
    \label{tab:components}
\end{table}

%% file: tables/ablation_Pra.tex
\begin{table}
    \centering
    \begin{tabular}{@{}cccccc@{}}
        \toprule
        MA& CA& SAC& AuPRC& FPR$_{95}$& mIoU\\
        \midrule
        \ding{51}& & & 81.24& 0.67& 58.03\\
        \ding{51}& \ding{51}& & 81.98& 0.53& 62.48\\
        \ding{51}& \ding{51}& \ding{51}& \textbf{85.56}& \textbf{0.45}& \textbf{63.30}\\
        \bottomrule
    \end{tabular}
    \caption{\textbf{Structural ablation of Prompt-based Restoration Attention.} MA: Masked Attention; CA: Cross Attention; SAC: Self-adaptive Correction.}
    \label{tab:pra}
    \vskip -4ex
\end{table}

%% file: tables/ablation_Bpdl.tex
\begin{table}[!t]
    \centering
    \resizebox{.9\linewidth}{!}{%
    \begin{tabular}{@{}ccccccc@{}}
        \toprule
        $\mathcal{L}_{pixel}$& $\mathcal{L}_{ind}$& $\mathcal{L}_{inDir}$& $\mathcal{L}_{outlier}$& AuPRC& FPR$_{95}$& mIoU\\
        \midrule
        \ding{51}& & & & 81.72& 0.54& 62.69\\
        \ding{51}& & \ding{51}& \ding{51}& 82.61& 0.50& 62.67\\
        \ding{51}& \ding{51}& & \ding{51}& 83.61& 0.49&63.04\\
        \ding{51}& \ding{51}& \ding{51}& & 83.37& 0.49& 63.06\\
        & \ding{51}& \ding{51}& \ding{51}& 83.08& 0.50& 62.89\\
        \ding{51}& \ding{51}& \ding{51}& \ding{51}& \textbf{85.56}& \textbf{0.45}& \textbf{63.30}\\
        \bottomrule
    \end{tabular}}
    \caption{\textbf{Ablation study of different loss contributions} in bilevel prompt distribution learning loss $\mathcal{L}_{bpdl}$.}
    \label{tab:ablationBpdl}
    \vskip -4ex
\end{table}

%% file: tables/ablation_OeFinetuning.tex
\begin{table*}[!t]
\centering
\setlength{\tabcolsep}{4pt}

\subfloat[\scriptsize Number of iterations
\label{tab:ablationIter}]{
    \begin{minipage}[b]{0.22\textwidth}
    \centering
    \resizebox{\linewidth}{!}{
        \begin{tabular}{@{}cccc@{}}
        \toprule
        Num Iter & AuPRC & FPR$_{95}$ & mIoU\\
        \midrule
        15$k$ & 79.19 & 0.49 & 57.49\\
        20$k$ & 78.22 & 0.57 & 60.14\\
        25$k$ & 84.04 & 0.51 & 62.71\\
        30$k$ & \textbf{85.56} & \textbf{0.45} & \textbf{63.30}\\
        35$k$ & 82.08 & 0.56 & 63.14\\
        \bottomrule
        \end{tabular}
    }
    \end{minipage}
}
\hfill
\subfloat[\scriptsize Outlier selection probability
\label{tab:ablationPout}]{
    \begin{minipage}[b]{0.2\textwidth}
    \centering
    \resizebox{\linewidth}{!}{
        \begin{tabular}{@{}cccc@{}}
        \toprule
        $p_{\mathrm{out}}$ & AuPRC & FPR$_{95}$ & mIoU\\
        \midrule
        0.1 & \textbf{86.30} & 0.45 & 62.47\\
        0.2 & 85.72 & 0.45 & 62.73\\
        0.3 & 85.56 & \textbf{0.45} & \textbf{63.30}\\
        0.4 & 81.28 & 0.59 & 62.17\\
        0.5 & 80.97 & 0.53 & 62.36\\
        \bottomrule
        \end{tabular}
    }
    \end{minipage}
}
\hfill
\subfloat[\scriptsize Different fine-tuning strategies
\label{tab:ablationStr}]{
    \begin{minipage}[b]{0.52\textwidth}
    \centering
    \resizebox{\linewidth}{!}{
        \begin{tabular}{@{}lccccccc@{}}
        \toprule
        Module
        & AuPRC
        & FPR$_{95}$
        & mIoU
        & NLL $\downarrow$
        & ECE (\%) $\downarrow$
        & mIoU$_{\mathrm{DP}}$
        & Params (M)\\
        \midrule
        \emph{w/o}
        & 55.66
        & 7.44
        & 54.11
        & 0.51
        & 5.70
        & 56.21
        & 0.40\\

        Decoders
        & 84.25
        & 0.50
        & 62.38
        & 0.39
        & 3.84
        & 64.81
        & 9.49\\

        Pixel Dec.
        & \textbf{85.56}
        & \textbf{0.45}
        & \textbf{63.30}
        & \textbf{0.38}
        & \textbf{3.81}
        & \textbf{66.07}
        & 7.07\\

        Trans. Dec.
        & 46.91
        & 5.58
        & 57.89
        & 0.48
        & 4.74
        & 60.99
        & 2.82\\
        \bottomrule
        \end{tabular}
    }
    \end{minipage}
}

\caption{
\textbf{Ablation study on outlier exposure fine-tuning.} 
Lower NLL and ECE indicate better probabilistic prediction quality and calibration, respectively. Params denotes the number of parameters updated during OE fine-tuning.}
\label{tab:oeFinetuning}
\vspace{-2ex}
\end{table*}

%% file: Tex_content/conclusion.tex
\section{Conclusion}
\label{sec:conclusion}
In this work, we introduce a novel task, termed \emph{Panoramic Out-of-Distribution Segmentation (PanOoS)}, which aims to enhance panoramic holistic scene understanding. 
To address \emph{PanOoS}, we propose the first panoramic out-of-distribution segmentation solution, POS, which effectively tackles the challenges posed by panoramas, including pixel distortions and background clutter. The proposed PRA optimizes the interaction between visual features and text embeddings, improving semantic understanding in complex panoramic scenes and mitigating semantic confusion. The BPDL further strengthens the model’s ability to recognize out-of-distribution regions by optimizing the embedding space in a hierarchical manner. The synergistic combination of PRA and BPDL allows POS to effectively resolve issues related to semantic ambiguity and geometric distortion in panoramic images. Moreover, we establish DenseOoS and QuadOoS benchmarks to facilitate the optimization and evaluation of \emph{PanOoS} models, paving the way for future research in \emph{PanOoS}. Extensive experiments on these two benchmarks demonstrate that POS achieves state-of-the-art performance in \emph{PanOoS}, significantly improving segmentation accuracy, robustness, and the detection of out-of-distribution regions. 

Despite its strong overall performance, POS exhibits reduced detection performance for outliers with extremely small apparent sizes, particularly under severe motion blur, defocus blur, or extreme illumination. Outliers whose appearances resemble those of in-distribution categories may also receive weak or incomplete responses. 
Its real-world generalization may further benefit from more diverse and realistic OE data. Future work will investigate improved OE construction, larger-scale real-world panoramic benchmarks, and temporal and multimodal cues for robust panoramic OoD perception.

%% file: Tex_content/appendix.tex
\clearpage
\appendices
\counterwithin{figure}{section}
\counterwithin{table}{section}

\section{Methodology}
\subsection{Prompt Templates}
\label{ap:promptTemplates}
We adopt the prompt templates from~\cite{radford2021learning,yu2023convolutions}, but further optimize them to more accurately describe items from different distributions. Specifically, the class and distribution prompt templates are as follows:

\subsubsection{Class}
\begin{quote}
    \textit{``A photo of a \{\}."\\
    ``This is a photo of a \{\}."\\
    ``There is a \{\} in the scene."\\
    ``There is the \{\} in the scene."\\
    ``A photo of a \{\} in the scene."\\
    ``A photo of a small \{\}."\\
    ``A photo of a medium \{\}."\\
    ``A photo of a large \{\}."\\
    ``This is a photo of a small \{\}."\\
    ``This is a photo of a medium \{\}."\\
    ``This is a photo of a large \{\}."\\
    ``There is a small \{\} in the scene."\\
    ``There is a medium \{\} in the scene."\\
    ``There is a large \{\} in the scene."}
\end{quote}

\subsubsection{Distribution}
\begin{quote}
    \textit{``A photo of an \{\} item."\\
    ``This is a photo of an \{\} item."\\
    ``There is an \{\} item in the urban or rural scene."\\
    ``There is the \{\} item in the urban or rural scene."\\
    ``A photo of an \{\} item in the urban or rural scene."\\
    ``A photo of a small \{\} item."\\
    ``A photo of a medium \{\} item."\\
    ``A photo of a large \{\} item."\\
    ``This is a photo of a small \{\} item."\\
    ``This is a photo of a medium \{\} item."\\
    ``This is a photo of a large \{\} item."\\
    ``There is a small \{\} item in the urban or rural scene."\\
    ``There is a medium \{\} item in the urban or rural scene."\\
    ``There is a large \{\} item in the urban or rural scene."}
\end{quote}

\section{Benchmarks and Experiments}
\subsection{More Details of Datasets}
\label{ap:detailsDatasets}

\subsubsection{DenseOoS}
\input{tables/sup/denseoos_classesNum}

\begin{figure}
    \centering
    \includegraphics[width=\linewidth]{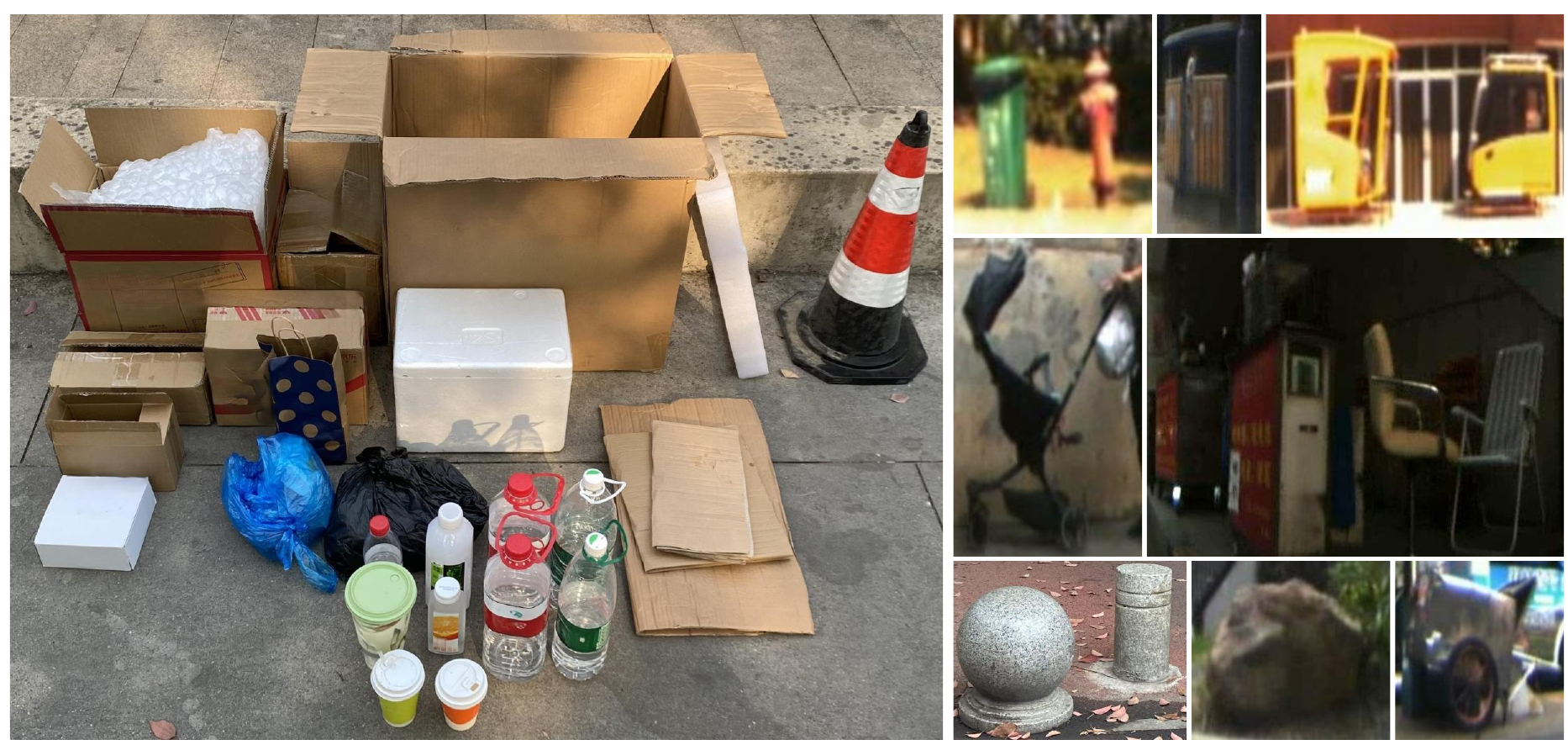}
    \caption{\textbf{Examples of outliers} included in the QuadOoS dataset. Including but not limited to the above types.}
    \label{fig:quadoosObjects}
\end{figure}

As shown in Table~\ref{tab:detailsDenseoos}, we present the specific types of outliers in DenseOoS. DenseOoS is based on the designs in POC~\cite{de2024placing} and additionally includes $5$ other common anomalous objects (\emph{cattle}, \emph{duck}, \emph{cock}, \emph{box}, and \emph{rock}). In addition, we also list some types of outliers in QuadOoS. As presented in Fig.~\ref{fig:quadoosObjects}, the outliers mainly consist of two parts: human-defined objects and outliers existing in the neighborhood.

\subsection{Evaluation Metrics}
\label{sec:evaluationMetrics}
In the context of the \emph{Panoramic Out-of-Distribution Segmentation (PanOoS)} benchmark, we employ three metrics, namely Area under PRC Curve (AuPRC), False Positive Rate at a true positive rate of $95\%$ (FPR$_{95}$), and mean Intersection-over-Union (mIoU), to evaluate the performance of models. For pixel-wise evaluation, $\mathcal{G}{\in}\{\mathcal{G}_{in},\mathcal{G}_{out}\}$ is the annotated ground truth labels for an image containing outliers, where $\mathcal{G}_{in}$ and $\mathcal{G}_{out}$ represent the inlier and outlier labels, respectively. 
Let $\mathcal{G}_{p}$ denote the model prediction obtained by a threshold $\tau$. The True Positive Rate (TPR), precision, and recall can be written as:
\begin{align}
    TPR(\tau)\!=\!recall(\tau)\!&=\!\frac{\mid\!\mathcal{G}_{p}\!\cap\!\mathcal{G}_{out}\!\mid}{\mid\!\mathcal{G}_{out}\!\mid}, \\
    precision(\tau)\!&=\!\frac{\mid\!\mathcal{G}_{p}\!\cap\!\mathcal{G}_{out}\!\mid}{\mid\!\mathcal{G}_{p}\!\mid}.
\end{align} 
The AuPRC can be approximated as:
\begin{align}
    AuPRC\!&=\!\int_{\tau}\!precision(\tau)\,recall(\tau).
\end{align}
The AuPRC works well for unbalanced datasets, which makes it particularly suitable for out-of-distribution segmentation since all the datasets are highly skewed. Next, we consider the FPR$_{95}$, an important criterion for safety-critical applications, which is calculated as:
\begin{equation}
    FPR_{95}\!=\!FPR(\tau_{95})\!=\!\frac{\mid\!\mathcal{G}_{p}(\tau_{95})\!\cap\!\mathcal{G}_{in}\!\mid}{\mid\!\mathcal{G}_{in}\!\mid},
\end{equation}
where $\tau_{95}$ is the threshold at which the TPR equals 95\%.

To effectively evaluate the in-distribution segmentation performance of the model, the standard mIoU is used for performance evaluation. The Intersection-over-Union (IoU) measures the overlap between predicted semantic segment $p$ and ground truth semantic segment $g$, while mIoU is the mean of IoU for each class in the dataset. It is calculated as:
\begin{equation}
    mIoU\!=\frac{1}{K}\!\sum_{k=1}^{K}\!\frac{p_{k}\!\cap\!g_{k}}{p_{k}\!\cup\!g_{k}}.
\end{equation}

To further evaluate the probabilistic quality of semantic predictions, we report Negative Log-Likelihood (NLL) and Expected Calibration Error (ECE). Both metrics are computed over the valid ID pixels in DenseOoS, excluding the OoD regions. Let $\Omega_{\mathrm{ID}}$ denote the set of valid ID pixels and $p_i(y_i)$ denote the predicted posterior probability of the ground-truth semantic class $y_i$ at pixel $i$. NLL is defined as
\begin{equation}
\mathrm{NLL}
=
-\frac{1}{|\Omega_{\mathrm{ID}}|}
\sum_{i\in\Omega_{\mathrm{ID}}}
\log p_i(y_i).
\end{equation}
A lower NLL indicates that the model assigns higher probabilities to the ground-truth semantic classes, with confident incorrect predictions receiving larger penalties.

For ECE, the predictions are partitioned into $M$ confidence bins $\{B_m\}_{m=1}^{M}$. It is computed as
\begin{equation}
\mathrm{ECE}
=
\sum_{m=1}^{M}
\frac{|B_m|}{|\Omega_{\mathrm{ID}}|}
\left|
\mathrm{acc}(B_m)-\mathrm{conf}(B_m)
\right|,
\end{equation}
where $\mathrm{acc}(B_m)$ and $\mathrm{conf}(B_m)$ denote the empirical accuracy and mean prediction confidence within the $m$-th bin, respectively. A lower ECE indicates better agreement between prediction confidence and semantic accuracy. Following~\cite{guo2017calibration}, we use $M=15$ equally spaced confidence bins.

\subsection{Comparative Experiments}
For the experiments on \emph{PanOoS}, we compare POS to the state-of-the-art pinhole-\emph{OoS} methods: per-pixel architectures~\cite{tian2022pixel,grcic2022densehybrid} and mask-transformer-based~\cite{grcic2023advantages,rai2024mask2anomaly,nayal2023rba} methods, using the models that were shared in their respective repositories.

\section{More Quantitative Results}
\label{appendix:more_quantitative_results}

\subsection{More In-Distribution Segmentation Results}
\input{tables/sup/panoSemanticIou}
To provide a thorough analysis of panoramic segmentation and demonstrate that strong closed-set segmentation performance ensures effective out-of-distribution segmentation, we present the accuracy results for each inlier class in the DenseOoS dataset, as shown in Table~\ref{tab:panoSemanticIou}. 
To benchmark our model, we conduct a comparative study with the existing state-of-the-art panoramic semantic segmentation methods, namely DAFormer~\cite{hoyer2022daformer}, Trans4PASS~\cite{zhang2022bending}, Trans4PASS+~\cite{zhang2024behind}, and 360SFUDA++~\cite{zheng2024360sfuda++}. 
DAFormer introduces domain-adaptive transformers to improve semantic segmentation across domains. Trans4PASS and its enhanced variant, Trans4PASS+, employ transformer-based architectures specifically designed to mitigate distortions and exploit contextual dependencies in panoramas. 360SFUDA++ represents an advanced source-free domain adaptation framework based on self-training, tailored to panoramic scenarios. 
While these methods achieve good performance under closed-set conditions, they experience degradation in the presence of out-of-distribution regions, owing to their limited generalization beyond the predefined semantic space. This limitation underscores the importance of enhancing robustness to unexpected content in real-world panoramic scenes.

As shown in the experimental results, our model achieves an mIoU of $63.30\%$, surpassing the performance of the previous best model by a large margin of $12.84\%$. 
A detailed breakdown of per-class performance reveals that our model excels across numerous categories, including \emph{road}, \emph{sidewalk}, \emph{building}, \emph{wall}, \emph{fence}, \emph{vegetation}, and \emph{terrain}. The performance in these classes demonstrates that our model is able to adapt well to the pixel distortions and background clutter introduced by the broader FoV of panoramic images, thus achieving leading segmentation performance.

\subsection{Additional Ablation Study}
\input{tables/ablation_TransDec}
\subsubsection{Features and Layers in Transformer Decoder}
To further analyze the most favorable settings for \emph{PanOoS}, we conduct additional ablation experiments on the decoded features and the corresponding number of layers in the transformer decoder. 
As shown in Table~\ref{tab:features2Transf}, the results align with previous work: decoding only semantically rich features and reducing the number of decoder layers alleviates semantic confusion, which helps to better align the logit scores and results in better outlier performance~\cite{nayal2023rba}. Specialized object queries perform worse with more decoder layers. 
This also supports our view that, while ensuring semantic clarity, enhancing the model's perception of panoramic pixels can improve the performance of \emph{PanOoS}.

\subsection{Efficiency Analysis}
\input{tables/sup/EfficiencyAnalysis}

To evaluate the deployment feasibility of POS in robotic and autonomous perception scenarios, we report its model parameters, FLOPs, inference speed, and peak GPU memory usage in Table~\ref{tab:efficiency_analysis}. POS has a computational profile of $361.27$M parameters and $3.25$ TFLOPs. During inference, it runs at $12.92$ FPS with a peak GPU memory usage of $4.21$GB. Considering its performance of $85.56\%$ AuPRC, $0.45\%$ FPR$_{95}$, and $63.30\%$ mIoU on DenseOoS, these results suggest that POS supports safety-aware out-of-distribution segmentation while maintaining practical computational efficiency.

\input{tables/sup/sizeAnalysis}

\subsection{Performance across Outlier Instance Sizes}
To quantify the effect of outlier size, we independently divide the GT OoD instances in each dataset according to their relative areas in the full panoramic image. Instances in the bottom $25\%$, middle $50\%$, and top $25\%$ are assigned to the \emph{Small}, \emph{Medium}, and \emph{Large} groups, respectively. Each connected GT OoD component in DenseOoS and QuadOoS is treated as an individual instance.

For each size group, AuPRC and FPR$_{95}$ are computed by treating the OoD pixels belonging to that group as positives and all valid ID pixels in the corresponding dataset as negatives. OoD pixels belonging to the other size groups are excluded from the group-specific evaluation. We additionally report macro-averaged per-instance recall at the single dataset-level threshold yielding $95\%$ overall OoD pixel recall. This measure assigns equal weight to each instance and therefore reduces the dominance of large objects in pixel-aggregated evaluation.

As shown in Table~\ref{tab:sizeAna}, POS achieves a $96.22\%$ macro-averaged per-instance recall for the \emph{Small} group on DenseOoS, whose median instance occupies only $0.142\%$ of the panorama, while maintaining an FPR$_{95}$ of $0.35\%$. 
However, a clearer size-related limitation is observed on the real-world QuadOoS benchmark. For the \emph{Small} group, the median instance area occupies only $0.014\%$ of the panorama, corresponding to merely $1.4$ out of every $10{,}000$ image pixels. Under this highly challenging setting, POS achieves a macro-averaged per-instance recall of $74.41\%$, an AuPRC of $0.16\%$, and an FPR$_{95}$ of $97.18\%$. As the median instance area increases to $0.061\%$ for the \emph{Medium} group and $0.289\%$ for the \emph{Large} group, the instance recall increases consistently from $74.41\%$ to $88.10\%$ and $94.65\%$, respectively. Meanwhile, AuPRC increases from $0.16\%$ to $10.38\%$ and $48.30\%$, whereas FPR$_{95}$ decreases from $97.18\%$ to $90.51\%$ and $81.10\%$. These results indicate that POS can detect a substantial proportion of small real-world outliers, but instances occupying extremely small image regions remain challenging under unconstrained panoramic conditions.

\input{tables/sup/categoryAnalysis}

\subsection{Category-wise Performance}
We further group the DenseOoS outliers into the five super-categories defined by the benchmark: \emph{Animals}, \emph{Garbage}, \emph{Furniture}, \emph{Vehicles} \& \emph{Tools}, and \emph{Others}. For each category, the pixel-level metrics are computed using all valid ID and OoD pixels from images belonging to that category.
As shown in Table~\ref{tab:categoryAna}, POS achieves an AuPRC of at least $80.21\%$ and an FPR$_{95}$ of at most $0.64\%$ across all five super-categories, indicating that its overall performance is not dominated by a single outlier type. Nevertheless, \emph{Vehicles} \& \emph{Tools} and \emph{Garbage} obtain the lowest instance recalls, at $89.40\%$ and $91.24\%$, respectively. Notably, \emph{Vehicles} \& \emph{Tools} has the largest median area but the lowest instance recall, indicating that its difficulty cannot be explained by object size alone. This result suggests that visual similarity to in-distribution traffic participants and roadside objects may constitute an additional source of ambiguity.

\section{More Qualitative Results}
\label{appendix:apd_more_qualitative_results}
\begin{figure*}
    \centering
    \includegraphics[width=\linewidth]{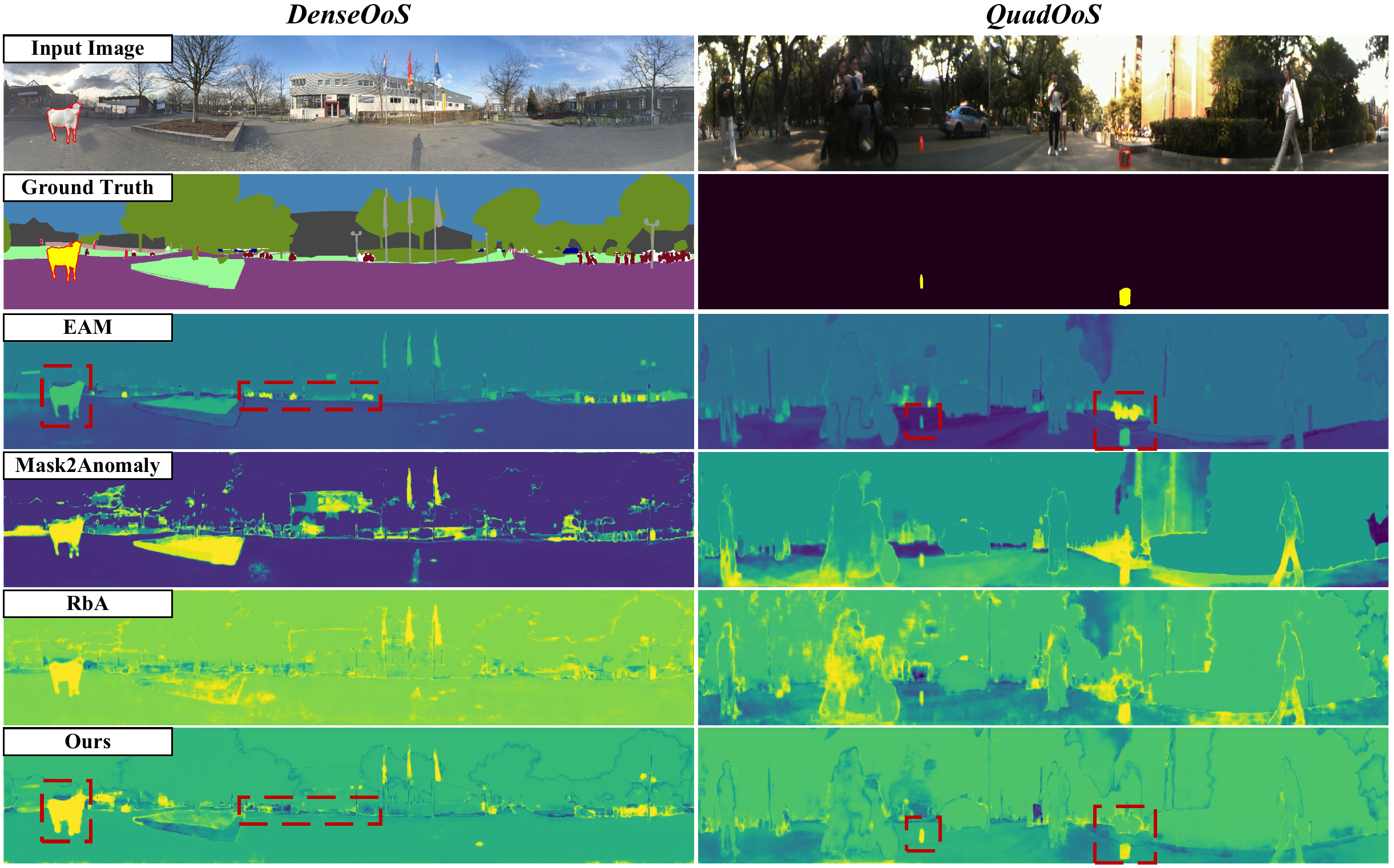}
    \caption{\textbf{Qualitative comparisons of \emph{PanOoS}.} 
    When existing pinhole-\emph{OoS} methods are transferred to \emph{PanOoS}, they tend to misidentify complex background regions as outliers. 
    Although EAM~\cite{grcic2023advantages} has a low false positive rate, it incorrectly identifies bicycles as outliers and nearly overlooks the actual outliers. In contrast, Mask2Anomaly~\cite{rai2024mask2anomaly} and RbA~\cite{nayal2023rba} suffer from significant false positives in the background regions, whereas POS delivers more precise \emph{PanOoS} across the $360^{\circ}$.}
    \label{fig:vis}
\end{figure*}

In Fig.~\ref{fig:vis}, we visualize the \emph{PanOoS} results produced by POS and other mask-transformer-based methods. 
Notably, POS effectively suppresses false positives in background regions, especially at the boundaries separating inlier regions, while better preserving the smoothness of the outlier map compared to other methods, despite sharing the same mask classification training paradigm. 

\begin{figure*}[t]
    \centering
    \includegraphics[width=\textwidth]{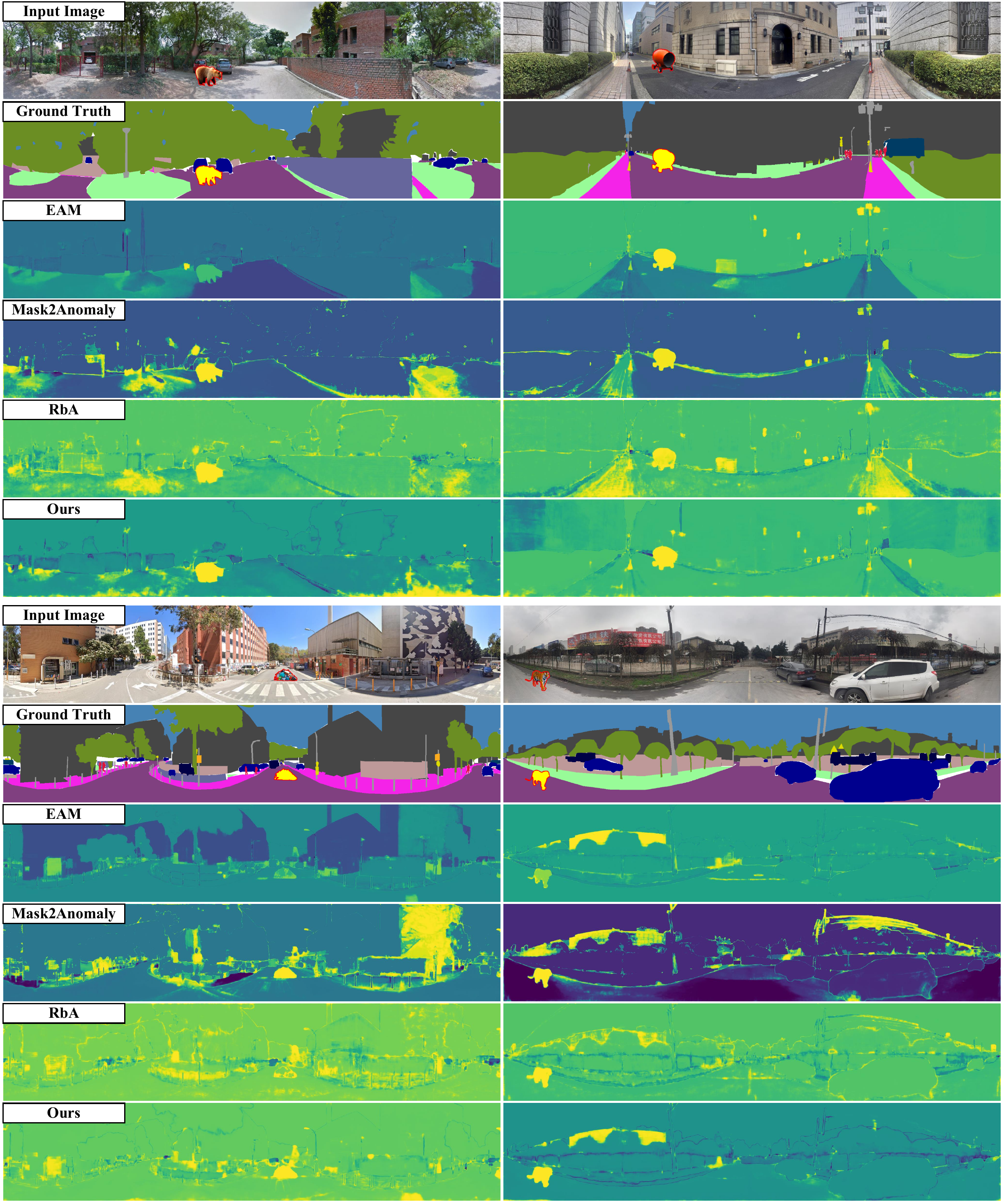}
    \caption{\textbf{Visualization results of \emph{PanOoS} on DenseOoS.} POS better distinguishes outliers from inliers and produces smoother outlier maps with fewer false positives. Zoom in for a better view.}
    \label{fig:denseoosResults}
\end{figure*}

\begin{figure*}[t]
    \centering
    \includegraphics[width=\textwidth]{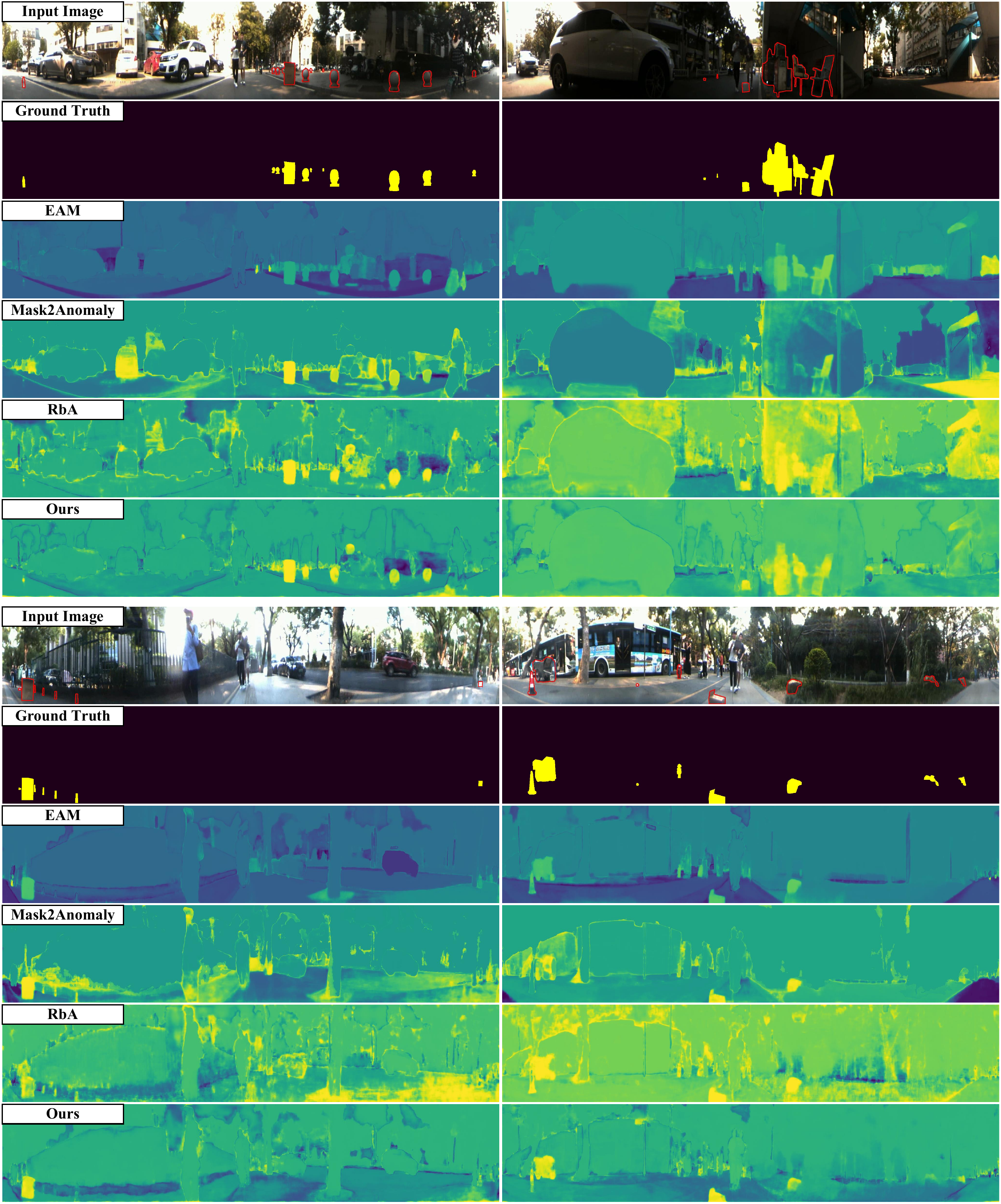}
    \caption{\textbf{Visualization results of \emph{PanOoS} on QuadOoS.} Uneven exposure, defocus, and motion blur caused by dense motion pose extra challenges to all methods. The pinhole-\emph{OoS} methods have serious omissions and false positives. However, POS achieves precise segmentation of outliers. Zoom in for a better view.}
    \label{fig:quadoosResults}
\end{figure*}

We showcase more visualization results of panoramic out-of-distribution segmentation of POS on DenseOoS and QuadOoS in Figs.~\ref{fig:denseoosResults} and~\ref{fig:quadoosResults}. 
Compared to other mask-transformer-based methods, EAM~\cite{grcic2023advantages}, Mask2Anomaly~\cite{rai2024mask2anomaly}, and RbA~\cite{nayal2023rba}, POS reduces false positives on the ambiguous background regions and boundaries of inliers while remaining sensitive to outliers. These improvements can be observed more prominently on QuadOoS (see Fig.~\ref{fig:quadoosResults}) under defocus and motion blur scenarios.

\section{Discussion}
\label{appendix:discussion}

\subsection{Societal Impacts}
In this study, we have introduced a novel task called \textbf{\emph{Panoramic Out-of-Distribution Segmentation (PanOoS)}} and established a comprehensive benchmark incorporating various well-known baseline models. 
We found that these baseline models exhibit limited performance on the \emph{PanOoS} task, primarily due to the pixel distortions and background clutter introduced by the panoramic broader FoV. 
To address this issue, we have developed POS, a solution that significantly enhances the performance on the \emph{PanOoS} task, outperforming existing pinhole-\emph{OoS} methods: pixel-wise architectures and mask-transformer-based methods, and achieving state-of-the-art results.

\emph{PanOoS} promotes more comprehensive omnidirectional scene understanding. Our work has the potential to support future anomaly detection applications for assisting people with visual impairments via robotic guide dogs and wearable robotics. However, it is also possible that it can be applied in the military field, such as for drone omnidirectional perception and autonomous strikes, endangering human safety. 
In addition, considering that the reliability of deep learning systems for advanced driver assistance systems is crucial, it is important to note that POS may still encounter misclassifications in challenging, intense-motion scenarios, potentially leading to erroneous predictions with adverse societal implications.

\begin{figure*}
    \centering
    \includegraphics[width=\textwidth]{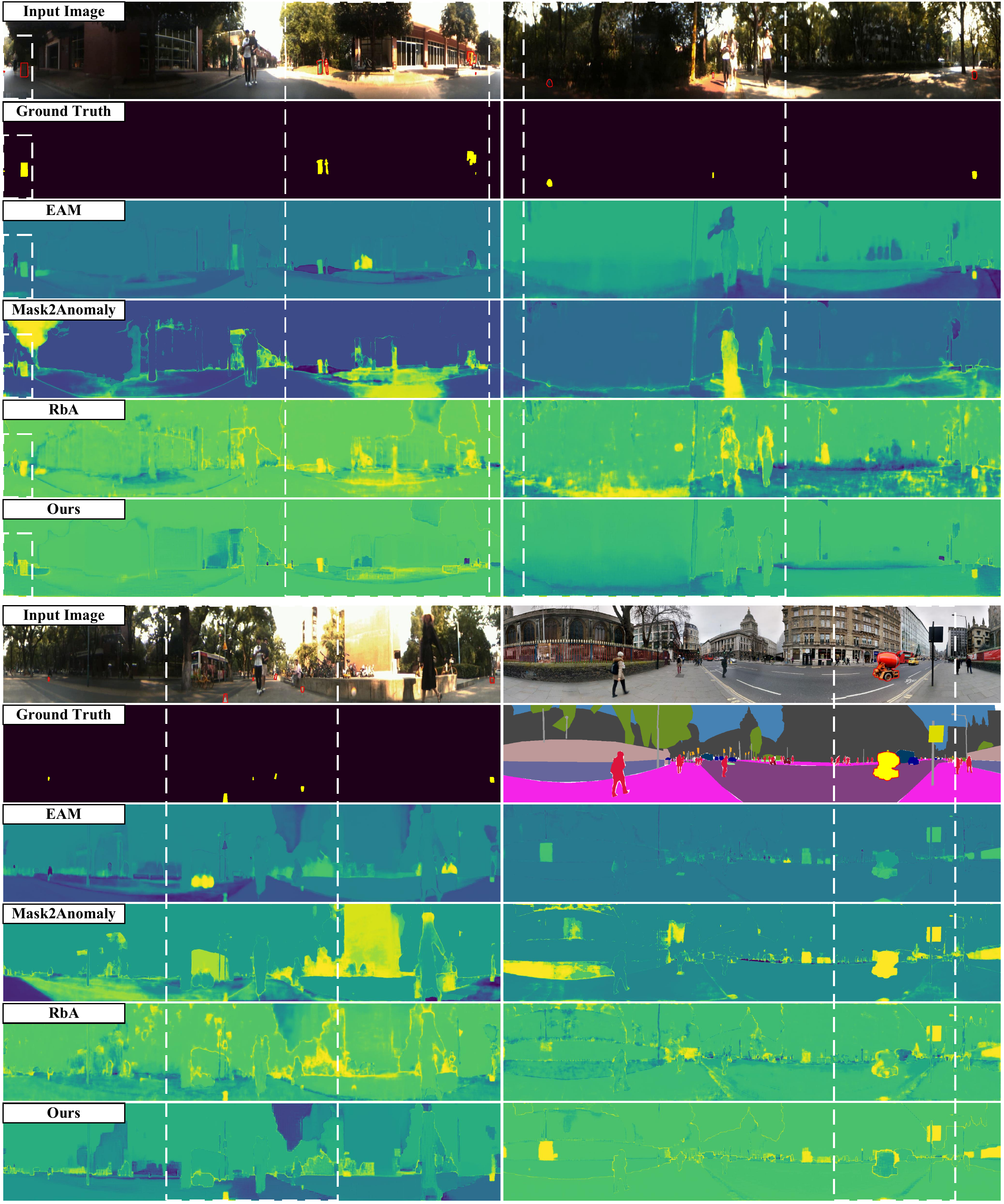}
    \caption{\textbf{Failure case analysis in \emph{PanOoS}.} Brighter colors indicate higher OoD confidence. The white dashed boxes highlight challenging regions, while the red solid contours indicate anomalous objects or regions.}
    \label{fig:failureCases}
\end{figure*}

\subsection{Failure Case Analysis}
Figure~\ref{fig:failureCases} compares the proposed POS with EAM~\cite{grcic2023advantages}, Mask2Anomaly~\cite{rai2024mask2anomaly}, and RbA~\cite{nayal2023rba}. In the shown examples, the compared methods produce widespread anomaly responses over several ID regions, whereas POS suppresses many of these background responses and produces more localized predictions.
Nevertheless, POS may assign weak responses to extremely small or distant outliers, particularly under low-light or overexposed conditions. In the final example, the cement mixer is only partially detected because its color, shape, and structural appearance resemble those of in-distribution vehicles. These observations are consistent with the quantitative size and category analyses and highlight small apparent size, extreme illumination, and appearance similarity to known categories as important observed failure modes.

\subsection{Limitations and Future Work}
Although POS effectively mitigates panoramic distortions and background clutter, its performance degrades under severe defocus blur, motion blur, and extreme illumination, particularly when outliers occupy very small image regions. Outliers that visually resemble in-distribution traffic participants or roadside objects may also produce weak or spatially incomplete responses. These observations indicate that panoramic OoD perception remains challenging when local appearance cues are degraded or semantically ambiguous.

The OoD score analysis further reveals a remaining gap between the controlled synthetic setting of DenseOoS and the unconstrained real-world conditions of QuadOoS, where OoD scores are more dispersed and exhibit increased overlap with ID scores. This discrepancy reflects differences in outlier appearance, imaging conditions, background complexity, and object scale. The real-world generalization of POS may also be influenced by the diversity and realism of the outlier-exposure data. Although the distinct sources used for training and evaluation provide evidence of cross-source generalization, the current experiments do not isolate the influence of OE quality.

Future work will investigate temporal and multimodal modeling, including depth and point-cloud cues, to improve robustness to small, blurred, partially occluded, or low-contrast outliers. More diverse and realistic OE construction will also be studied through controlled comparisons of different exposure strategies. These directions may further improve \emph{PanOoS} performance and support extensions to open-set and 3D scene perception.
In parallel, more realistic evaluation benchmarks could be developed by incorporating depth-aware placement and occlusion modeling, illumination and shadow matching, and panoramic-specific blur, exposure, and distortion simulation into synthetic outlier generation.
Expanding QuadOoS into a larger-scale robotic benchmark for \emph{PanOoS} remains an important direction for future benchmark-side research. Future extensions could be built by collecting more real-world panoramic images from quadruped robots and other mobile platforms across diverse indoor and outdoor scenarios, covering richer scene types, lighting conditions, weather variations, motion patterns, viewpoints, and outlier categories. To better reflect real-world deployment challenges, future benchmarks should further consider factors such as motion blur, defocus, uneven exposure, color inconsistency, and dynamic background interference. Standardized annotations and evaluation protocols would also be important for enabling more comprehensive and stable evaluation of \emph{PanOoS} in real-world robotic environments.

%% file: tables/sup/denseoos_classesNum.tex
\begin{table}
    \centering
    \begin{tabular}{@{}lcc@{}}
        \toprule
        Class & ID & Number  \\
        \midrule    
        stroller
        &  19
        &  19
        \\
        trolley
        &  20
        &  20
        \\
        garbage bag
        &  21
        &  55
        \\
        wheelie bin
        &  22
        &  41
        \\
        suitcase
        &  23
        &  16
        \\
        skateboard
        &  24
        &  13
        \\
        chair dumped on the street
        &  25
        &  23
        \\
        sofa dumped on the street
        &  26
        &  41
        \\
        furniture dumped on the street
        &  27
        &  32
        \\
        mattress dumped on the street
        &  28
        &  63
        \\
        garbage dumped on the street
        &  29
        &  90
        \\
        clothes dumped on the street
        &  30
        &  101
        \\
        cement mixer on the street
        &  31
        &  48
        \\
        cat
        &  32
        &  10
        \\
        dog
        &  33
        &  10
        \\
        bird flying
        &  34
        &  10
        \\
        horse
        &  35
        &  34
        \\
        skunk
        &  36
        &  38
        \\
        sheep
        &  37
        &  48
        \\
        crocodile
        &  38
        &  19
        \\
        alligator
        &  39
        &  12
        \\
        bear
        &  40
        &  65
        \\
        llama
        &  41
        &  18
        \\
        tiger
        &  42
        &  38
        \\
        monkey
        &  43
        &  7
        \\
        cattle
        &  44
        &  43
        \\
        duck
        &  45
        &  28
        \\
        cock
        &  46
        &  12
        \\
        box
        &  47
        &  25
        \\
        rock
        &  48
        &  21
        \\
        \textbf{total}
        &  -&  1000\\
        \bottomrule
    \end{tabular}
    \caption{\textbf{Details of the DenseOoS dataset.}}
    \label{tab:detailsDenseoos}
\end{table}

%% file: tables/sup/panoSemanticIou.tex
\begin{table*}[!t]
    \centering
    \resizebox{\textwidth}{!}{%
    \begin{tabular}{@{}lccccccccccccccccccccc@{}}
        \toprule
        Method& road&  sidewalk& building& wall& fence & pole& traffic light& traffic sign& vegetation& terrain& sky& person& rider& car& truck&bus & train& motorcycle& bicycle& Metric\\
        \midrule    
        DAFormer~\cite{hoyer2022daformer}& 48.94& 31.78& 84.12& 26.51& 31.06& 24.92& 6.83& 7.73& 71.86& 40.63& 91.23& 44.13& 2.1& 78.02& 55.02& 25.24& 84.89& 61.37& 29.59& 44.53\\
        Trans4PASS (S)~\cite{zhang2022bending}& 74.28& 30.68& 84.79& 28.45& 36.60& 27.90& 15.40& 15.11& 78.99& 33.71& 93.28& 49.69& 6.79& 82.00& 46.42& 51.79& 72.84& 68.34& 40.63&49.35\\
        Trans4PASS+ (S)~\cite{zhang2024behind}& 74.74& 34.17& 85.43& 23.46& 42.78& 28.45& 17.14& 18.08& 79.64& 29.41& 93.00& 48.74& 5.40& 82.76& 50.54& 44.91& 91.49& 61.49& 40.20&50.11\\
        360SFUDA++~\cite{zheng2024360sfuda++}& 72.29& 45.24& 84.91& 32.11& 38.18& 29.83& 19.28& 17.98& 76.02& 27.23& 92.01& 53.14& \textbf{18.74}& 81.35& 44.42& 45.38& 82.84& 60.93& 36.92&50.46\\
        Ours& \textbf{87.67}&  \textbf{59.56}& \textbf{89.68}& \textbf{50.88}&  \textbf{45.10}& \textbf{42.62}& \textbf{34.02}& \textbf{22.73}& \textbf{81.21}& \textbf{45.19}& \textbf{94.59}& \textbf{72.24}& 14.42& \textbf{86.62}& \textbf{64.88}& \textbf{69.16}& \textbf{95.27}& \textbf{82.98}& \textbf{63.92}& \textbf{63.30}\\
        \bottomrule
    \end{tabular}}
    \caption{\textbf{Panoramic semantic segmentation results} on the DenseOoS benchmark. The per-class results are reported as IoU, and the metric is mIoU.}
    \label{tab:panoSemanticIou}
\end{table*}

%% file: tables/ablation_TransDec.tex
\begin{table}
    \centering
    \begin{tabular}{@{}ccccc@{}}
        \toprule
        Features & Layer & AuPRC & FPR$_{95}$ & mIoU \\
        \midrule    
        \multirow{3}{*}{$f_{4}$} & 1& \textbf{85.56}& \textbf{0.45}& \textbf{63.30}\\
        & 3& 71.18& 0.68& 57.72\\
        & 6& 76.46& 0.64& 57.33\\
        \hline
        \multirow{3}{*}{$f_{2}$, $f_{3}$, $f_{4}$} & 3& 72.38& 0.94& 58.96\\
        & 6& 63.60& 1.01& 58.66\\
        & 9& 65.91& 1.06& 58.79\\
        \bottomrule
    \end{tabular}
    \caption{\textbf{Comparison of different features in Transformer Decoder.} The $f_{2}$ and $f_{3}$ are feature maps with stride $8(c_{8})$ and $16(c_{16})$ processed by DeformAttn layers, respectively.}
    \label{tab:features2Transf}
\end{table}

%% file: tables/sup/EfficiencyAnalysis.tex
\begin{table*}[ht]
\centering
\resizebox{\linewidth}{!}{
\begin{tabular}{lccccccccc}
    \toprule
    \multirow{2}{*}{Method}
    & \multicolumn{6}{c}{Efficiency Metrics}
    & \multicolumn{3}{c}{DenseOoS} \\
    \cmidrule(lr){2-7} \cmidrule(lr){8-10}
    & Input Res.
    & Total Params (M) $\downarrow$
    & Trainable Params (M) $\downarrow$
    & TFLOPs $\downarrow$
    & FPS $\uparrow$
    & Mem. (GB) $\downarrow$
    & AuPRC $\uparrow$
    & FPR95 $\downarrow$
    & mIoU $\uparrow$ \\
    \midrule
    \textbf{Ours}
    & $400{\times}2048$
    & 361.27
    & 9.49
    & 3.25
    & 12.92
    & 4.21
    & 85.56
    & 0.45
    & 63.30 \\
    \bottomrule
\end{tabular}}
\caption{\textbf{Efficiency analysis of POS when inferring with RTX 3090 on DenseOoS images.}}
\label{tab:efficiency_analysis}
\end{table*}

%% file: tables/sup/sizeAnalysis.tex
\begin{table}[!t]
    \centering
    \resizebox{\linewidth}{!}{
    \begin{tabular}{llccccc}
        \toprule
        Dataset
        & Size
        & Inst.
        & Median Area (\%)
        & AuPRC
        & FPR$_{95}$
        & Inst. Recall~$\uparrow$ \\
        \midrule
        \multirow{4}{*}{DenseOoS}
        & Small
        & 250
        & 0.142
        & 50.75
        & 0.35
        & 96.22 \\
        
        & Medium
        & 500
        & 0.353
        & 77.05
        & 0.43
        & 95.16 \\
        
        & Large
        & 250
        & 0.743
        & 76.56
        & 0.49
        & 94.93 \\
        
        & \textbf{Overall}
        & 1000
        & 0.353
        & 85.56
        & 0.45
        & 95.37 \\
        \midrule
        
        \multirow{4}{*}{QuadOoS}
        & Small
        & 184
        & 0.014
        & 0.16
        & 97.18
        & 74.41 \\
        
        & Medium
        & 367
        & 0.061
        & 10.38
        & 90.51
        & 88.10 \\
        
        & Large
        & 184
        & 0.289
        & 48.30
        & 81.10
        & 94.65 \\
        
        & \textbf{Overall}
        & 735
        & 0.061
        & 48.13
        & 85.22
        & 86.31 \\
        \bottomrule
    \end{tabular}
    }
    \caption{
    \textbf{Performance of POS across different outlier-instance sizes on DenseOoS and QuadOoS.} Instances are grouped independently for each dataset according to their relative GT areas: Small (bottom $25\%$), Medium (middle $50\%$), and Large (top $25\%$). Median Area denotes the median instance area relative to the full panoramic image. Inst. Recall is the macro-averaged per-instance recall at the dataset-level threshold corresponding to $95\%$ OoD pixel recall. Overall reports performance on the complete dataset.}
    \label{tab:sizeAna}
\end{table}

%% file: tables/sup/categoryAnalysis.tex
\begin{table}
    \centering
    \resizebox{\linewidth}{!}{
    \begin{tabular}{lccccc}
        \toprule
        Category
        & Images
        & Median Area (\%)
        & AuPRC
        & FPR$_{95}$
        & Inst. Recall~$\uparrow$ \\
        \midrule
        Animals
        & 392
        & 0.275
        & 85.65
        & 0.34
        & 97.97 \\
        
        Garbage
        & 145
        & 0.354
        & 80.21
        & 0.63
        & 91.24 \\
        
        Furniture
        & 159
        & 0.363
        & 90.65
        & 0.29
        & 95.76 \\
        
        Vehicles \& Tools
        & 141
        & 0.585
        & 83.62
        & 0.64
        & 89.40 \\
        
        Others
        & 163
        & 0.373
        & 92.47
        & 0.23
        & 97.58 \\
        
        \textbf{Overall}
        & 1000
        & 0.353
        & 85.56
        & 0.45
        & 95.37 \\
        \bottomrule
    \end{tabular}
    }
    \caption{\textbf{Category-wise OoD detection performance of POS on DenseOoS.} The five super-categories follow the DenseOoS taxonomy. Pixel-level metrics are computed using all valid ID and OoD pixels from images belonging to each category.}
    \label{tab:categoryAna}
\end{table}